\documentclass{article}

\usepackage{xcolor}
\usepackage[preprint]{corl_2026} % Uncomment for the camera-ready ``final'' version.
\usepackage{amsmath,amsfonts,bm}

\def\eqref#1{equation~\ref{#1}}
\def\1{\bm{1}}

\DeclareMathAlphabet{\mathsfit}{\encodingdefault}{\sfdefault}{m}{sl}
\SetMathAlphabet{\mathsfit}{bold}{\encodingdefault}{\sfdefault}{bx}{n}

\usepackage{hyperref}
\usepackage{url}
\usepackage{footmisc}
\usepackage{wrapfig}

\usepackage{amsmath}
\usepackage{amssymb}
\usepackage{mathtools}
\usepackage{amsthm}
\usepackage{tabularx}
\usepackage{utfsym}

\definecolor{ForestGreen}{rgb}{0.1333,0.545,0.1333}
\definecolor{Firebrick}{rgb}{0.698,0.1333,0.1333}
\newcommand{\Ours}{CAIP}
\usepackage{caption}

\usepackage{subcaption}
\usepackage{caption}
\usepackage{placeins}
\usepackage{xspace}
\usepackage{multirow}     
\usepackage[dvipsnames]{xcolor}

\definecolor{ForestGreen}{rgb}{0.1333,0.545,0.1333}
\definecolor{Firebrick}{rgb}{0.698,0.1333,0.1333}

\newcommand{\minisection}[1]{\noindent{\textbf{#1.}}}
\newcommand{\tablestyle}[2]{\setlength{\tabcolsep}{#1}\renewcommand{\arraystretch}{#2}\centering\footnotesize}
\newlength\savewidth

\usepackage{xcolor}  % Include the package

\definecolor{robotaction}{RGB}{255, 140, 0}  % Orange color
\definecolor{robottype}{RGB}{178, 34, 34}
\definecolor{robottask}{RGB}{65, 105, 225}  % Orange color
\definecolor{controltype}{RGB}{0. 205, 0}
\definecolor{predsteps}{RGB}{216, 191, 216}
\definecolor{lavendermist}{rgb}{0.9, 0.9, 0.98}
\definecolor{visualtrace}{RGB}{255, 216, 0} % Yellow
\usepackage{colortbl}
\usepackage{xcolor}
\definecolor{lightgray}{gray}{0.9}
\usepackage{booktabs}
\usepackage{color, soul}
\usepackage{graphicx}
\usepackage{amsmath}
\usepackage{empheq}
\usepackage{cleveref}
\usepackage{epigraph}
\definecolor{lightgray}{gray}{0.9}
\definecolor{lightblue}{rgb}{0.93,0.95,1.0}
\definecolor{darkgreen}{rgb}{0.0,0.6,0.0}
\definecolor{darkblue}{rgb}{0.0,0.0,0.5}
\definecolor{pinegreen}{rgb}{0.0, 0.47, 0.44}
\definecolor{deepmagenta}{rgb}{0.8, 0.0, 0.8}
\definecolor{amber}{rgb}{1.0, 0.49, 0.0}

\newcommand{\ignorebig}[1]{}

\definecolor{lightred}{RGB}{241,140,142}
\definecolor{amber(sae/ece)}{rgb}{1.0, 0.49, 0.0}
\definecolor{battleshipgrey}{rgb}{0.52, 0.52, 0.51}
\definecolor{cadmiumorange}{rgb}{0.93, 0.53, 0.18}
\definecolor{applegreen}{rgb}{0.55, 0.71, 0.0}
\definecolor{cadmiumgreen}{rgb}{0.0, 0.42, 0.24}
\definecolor{forestgreen}{rgb}{0.13, 0.55, 0.13}
\definecolor{red}{rgb}{0.89, 0.0, 0.13}

\definecolor{cb-0}{RGB}{216, 27, 96}
\definecolor{cb-1}{RGB}{30,136,229}
\definecolor{cb-2}{RGB}{255,193,7}
\definecolor{cb-3}{RGB}{0, 77, 64}
\definecolor{cb-4}{RGB}{150,220,174}

\title{Contrastive Action-Image Pre-training for Visuomotor Control}

\author{Yuvan Sharma$^{1,\ast}$, Dantong Niu$^{1,2,\ast, \ddagger}$, Anirudh Pai$^{1,\ast}$, Zekai Wang$^{1}$, Zhuoyang Liu$^{1}$, \\ 
\textbf{Baifeng Shi}$^{1}$\textbf{, Stefano Saravalle}$^{3}$\textbf{, Boning Shao}$^{1}$\textbf{,  Ruijie Zheng}$^{2}$\textbf{, Jing Wang}$^{2}$\textbf{,}\\
\textbf{Konstantinos Kallidromitis}$^{4}$\textbf{, Yusuke Kato}$^{4}$\textbf{, Fabio Galasso}$^{3,5}$\textbf{, Yuke Zhu}$^{2}$\textbf{, Danfei Xu}$^{2}$\textbf{,} \\ 
\textbf{Linxi ``Jim" Fan}$^{2}$\textbf{, Jitendra Malik}$^{1,\dagger}$\textbf{, Trevor Darrell}$^{1,\dagger}$\textbf{, Roei Herzig}$^{1,\dagger}$\\[1em] 
\textsuperscript{1}UC Berkeley \quad
\textsuperscript{2}NVIDIA \quad
\textsuperscript{3}Sapienza University of Rome\quad
\textsuperscript{4}Panasonic\quad
\textsuperscript{5}ItalAI\\[1em] 
\textsuperscript{*}Equal Contribution \quad
\textsuperscript{$\ddagger$}Project Lead \quad
\textsuperscript{$\dagger$}Equal Advising}

\begin{document}
\maketitle
%===============================================================================
% However, such action annotations are difficult to obtain, since robot data is expensive to collect at scale. 
% While obtaining this paired signal directly from robot trajectories is cost-prohibitive at scale, egocentric human videos offer an abundant alternative source of action data. 
% However, scaling this paired signal through robot trajectories is unfeasible due to the lack of robotic annotations, forcing a shift toward extracting action data from abundant human videos

\begin{abstract}
Existing vision encoders for robotics face a fundamental bottleneck: robotic datasets lack the scale necessary for large-scale pre-training. 
Prior work circumvents this data scarcity by turning to internet-scale image and language data or egocentric human video. 
While these models show promise, neither paradigm learns from paired vision and action data, which downstream visuomotor control policies require. 
However, robot trajectories, the most direct source of this paired signal, are not available at pre-training scale, motivating us to extract action signals from abundant human video instead. 
To this end, we introduce CAIP (Contrastive Action-Image Pre-training), a vision encoder that treats human hand poses from large-scale egocentric video as a proxy for end-effector actions.
By extracting 3D hand keypoints, a representation that aligns naturally with downstream robot action spaces, CAIP learns a unified action–image representation through a contrastive objective.
Leveraging 32,041 hours of egocentric human video and only 88 hours of robotic manipulation data, CAIP outperforms state-of-the-art vision encoders including DINOv2, SigLIP, MVP, and R3M. 
Evaluated on a challenging real-world dexterous manipulation setup using Dexmate Vega and Sharpa Wave hands, CAIP yields performance gains of more than 30\% on tasks involving folding, pouring, and fine-grained manipulation. Our results show that our method of contrastive action-centric pre-training yields a scalable path to achieving robust visual representations better suited for physical interaction.
\end{abstract}
\keywords{Action-Image Contrastive Learning, Dexterous Manipulation} 

%===============================================================================
\begin{figure}[t!] 
    \centering 
    \includegraphics[width=1.0\linewidth]{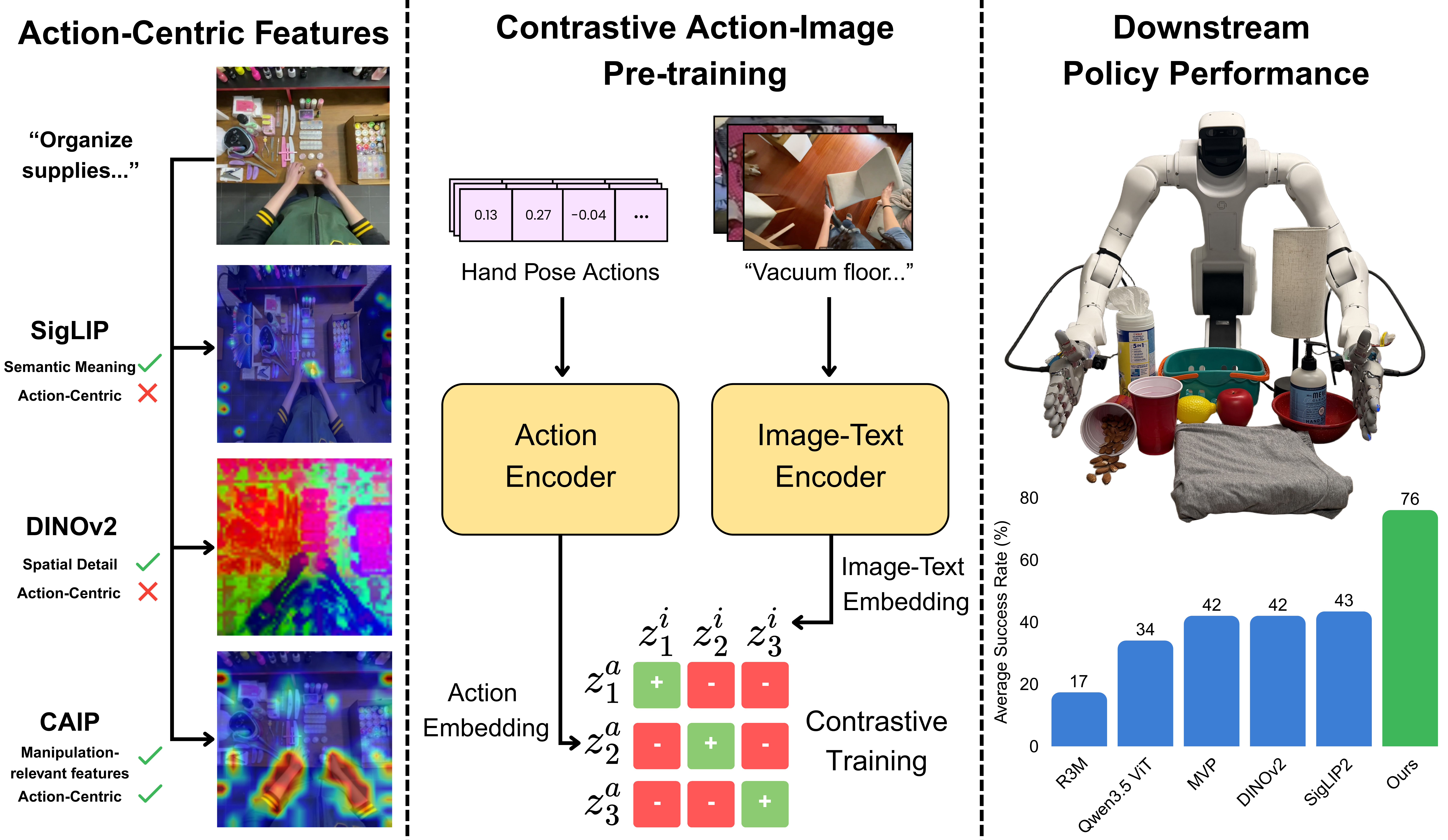} 
    \caption{
\textbf{(Left)} We visualize which image regions each encoder emphasizes, with saliency being computed using each encoder's natural query mechanism (see~\Cref{app:saliency}). SigLIP captures high-level semantics and DINOv2 captures visual structure, but neither attends to action-relevant regions. Our encoder produces manipulation-centric features focused on hands and relevant objects.
\textbf{(Center)} Hand pose actions and paired image-text inputs are encoded separately, then aligned via a SigLIP-style contrastive loss.
\textbf{(Right)} CAIP achieves superior performance on real-world tasks compared to state-of-the-art vision encoders such as SigLIP 2, DINOv2, and MVP (see~\Cref{sec:exp}).
}
    \label{fig:teaser} 
\end{figure}

\section{Introduction}

Visual perception is fundamental to robotic manipulation, as a robot’s ability to reason about its environment and perform precise interactions depends critically on the quality of its visual features. Dominant visual pre-training paradigms such as image-text contrastive learning~\citep{radford2021learningtransferablevisualmodels, zhai2023sigmoidlosslanguageimage}, self-supervised reconstruction~\citep{he2021maskedautoencodersscalablevision}, and self-distillation~\citep{oquab2024dinov2learningrobustvisual} have produced encoders with remarkable semantic and visual understanding. Although these backbones have driven major advances in vision~\cite{liu2023visualinstructiontuning, bao2022beitbertpretrainingimage} and language tasks~\citep{alayrac2022flamingovisuallanguagemodel, li2023blip2bootstrappinglanguageimagepretraining}, they were not designed with physical interaction in mind. For example, semantic encoders like CLIP~\cite{radford2021learningtransferablevisualmodels} and SigLIP~\cite{zhai2023sigmoidlosslanguageimage} provide high-level semantic knowledge, while encoders like DINO~\cite{oquab2024dinov2learningrobustvisual, caron2021emergingpropertiesselfsupervisedvision} capture fine-grained geometric details like depth and segmentation. 
However, neither class of encoder sees manipulation environments during training, nor receives direct action supervision. This creates a fundamental misalignment: current models have a strong semantic understanding of a scene (aligning language and vision), but lack the action-centric structure that we demonstrate benefits downstream policy learning.

The most direct way to close this gap is to pre-train on robot data itself, where action labels are natively available. In practice, however, robot trajectories are notoriously difficult to collect at scale. While recent large-scale datasets such as DROID~\citep{khazatsky2025droidlargescaleinthewildrobot} and Open X-Embodiment~\citep{embodimentcollaboration2025openxembodimentroboticlearning} have led to significant progress, the volume of available robot trajectories remains orders of magnitude smaller than internet-scale video corpora. This scarcity naturally motivates the search for alternative sources of action-rich data. Egocentric human video datasets~\citep{grauman2022ego4daroundworld3000, Shan20, Damen2018EPICKITCHENS} offer an abundant repository of human–object interactions, yet they lack the explicit robotic labels required for action-centric pre-training. Consequently, prior approaches such as R3M~\citep{nair2022r3muniversalvisualrepresentation} and MVP~\citep{xiao2022maskedvisualpretrainingmotor} resort to alternative objectives like frame-level contrastive loss or masked autoencoder reconstruction. While these objectives produce useful visual features, they omit the action-conditioned information that is critical for learning downstream control. In this work, we propose that human hand poses can serve as a powerful proxy for these missing robotic labels. By leveraging these poses in a form analogous to robotic end-effector actions, we bridge the gap between abundant human demonstrations and sparse robotic data to learn representations better suited for manipulation.

Building on this insight, we introduce \textbf{CAIP} (Contrastive Action-Image Pre-training), a vision encoder trained via a contrastive objective on large-scale egocentric video paired with extracted hand pose labels (see \Cref{fig:teaser}). We unify these diverse egocentric video sources into a shared action space and learn the relationship between image-text representations and their corresponding action signals. This formulation yields a modular vision encoder, and we empirically demonstrate that its action-centric representations enable more robust and capable policies. Crucially, CAIP achieves this with minimal robot data, leveraging abundant human video and using action supervision to structure the visual representation space.

We summarize our contributions as follows: (i) We propose a contrastive training methodology for learning action-centric visual representations from egocentric human video, directly grounded in explicit action labels. (ii) We curate and unify large-scale egocentric video datasets to train on 32,129 hours of manipulation data. (iii) We show that our learned encoder consistently outperforms current state-of-the-art vision encoders such as DINOv2, SigLIP, and MVP. (iv) We demonstrate that our learned representations generalize to out-of-distribution settings, enabling robust downstream policy performance under environmental variation.
\section{Contrastive Action-Image Pre-training}
We introduce \textit{CAIP}, an action-centric vision encoder that learns manipulation-relevant visual representations by aligning egocentric scenes with paired hand actions. Inspired by image-text contrastive methods~\cite{radford2021learningtransferablevisualmodels,zhai2023sigmoidlosslanguageimage}, CAIP optimizes a contrastive objective between a joint image–text latent space and an action latent space. We train CAIP on vast egocentric human data, exposing our model to the diverse embodiments and environments needed to learn robust visual representations.

\begin{figure}[t!] 
    \centering 
    \includegraphics[width=1.0\linewidth]{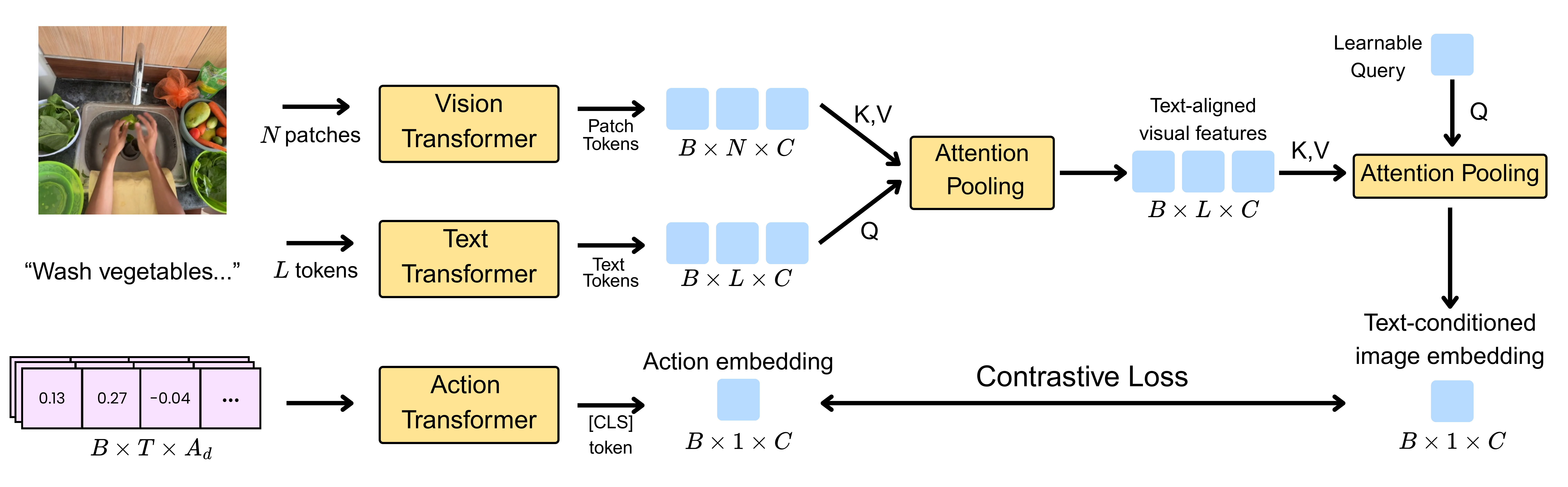} 
    \caption{\textbf{CAIP architecture.} A ViT encodes $N$ image patches and a text transformer encodes $L$ language tokens, while an action transformer encodes a $T$-step action chunk into a single embedding via the $[\text{CLS}]$ token. To form a text-conditioned image embedding, we attention-pool patch tokens using text tokens as queries, then pool the result with a learnable query. The action embedding and text-conditioned image embedding are aligned via a SigLIP contrastive loss.}
    \label{fig:architecture} 
\end{figure}

\subsection{Pre-training Architecture}
\label{sec:pretrain-arch}
We pre-train a vision encoder that produces text-conditioned image features that align with action features through a contrastive objective. The architecture consists of three encoders: one for each of vision, language, and action. Their outputs are combined through attention pooling before the contrastive loss.~\Cref{fig:architecture} illustrates the architecture. Our vision encoder uses a ViT-L/16 backbone for the image tower and a 24-layer transformer for the text tower, both initialized from SigLIP 2~\citep{tschannen2025siglip2multilingualvisionlanguage}.

\minisection{Vision encoder} An input image $I$ is split into $N$ patches and processed by a ViT backbone, yielding patch features of shape $B \times N \times C$, where $B$ is the batch size and $C$ is the embedding dimension. We do not perform global pooling at this stage; the full patch sequence is retained so that text tokens can attend to spatially localized features in the next step.

\minisection{Language encoder} The accompanying natural-language instruction is tokenized into $L$ tokens and passed through a transformer text encoder, producing token-level features of shape $B \times L \times C$.

\minisection{Action encoder} Each training sample includes a sequence of future actions of shape $B \times T \times A_d$, where $T$ is the prediction horizon and $A_d=378$ is the per-timestep action dimensionality ($378 = 42 \times 9$, corresponding to 21 keypoints per hand represented through 9D pose). A lightweight 4-layer transformer encoder processes this sequence, and the CLS token is extracted to produce a single action embedding of shape $B \times 1 \times C$. The action encoder is trained from scratch.

\minisection{Text-conditioned image pooling} To produce text-conditioned image features, we apply two stages of attention pooling. First, language tokens ($B \times L \times C$) serve as queries while image patch features ($B \times N \times C$) serve as keys and values, producing text-grounded visual features of shape $B \times L \times C$. Second, a learnable query token attends over these text-grounded features to produce a single text-conditioned image embedding of shape $B \times 1 \times C$.
\subsection{Pre-training Objective}
\label{sec:pretrain-objective}

We adopt a SigLIP-style sigmoid contrastive loss~\citep{zhai2023sigmoidlosslanguageimage} to align text-conditioned image embeddings with action embeddings. Unlike the softmax-based InfoNCE loss used in CLIP~\citep{radford2021learningtransferablevisualmodels}, the SigLIP objective treats each pair independently as a binary classification problem, which removes the need for a global normalization across the batch and improves training stability at large batch sizes. The loss is described in further detail in~\Cref{app:vision_training_details}.

\subsection{Downstream Policy}
\label{sec:downstream}

We evaluate the pre-trained vision encoder by transferring it to a closed-loop manipulation policy.

\minisection{Policy architecture}\label{method:downstream} The policy takes as input head and two wrist camera images and a natural-language task instruction, and outputs a chunk of future actions to be executed on the robot. Visual and language inputs are processed by the frozen pre-trained encoder, producing per-patch visual tokens (one set per camera view) and text tokens. Each modality is projected through a learnable linear layer into the policy's hidden dimension and interleaved into a single token sequence.

The token sequence is processed by a decoder-only transformer (Qwen3.5-0.8B~\citep{qwen3.5}). All weights are trained from scratch rather than initialized from the pre-trained VLM, providing a strong sequence-modeling architecture while enabling fair comparison with baselines.

\minisection{Action head} Conditioned on the backbone's output representation, the policy predicts the action chunk using a flow-matching objective~\citep{lipman2023flowmatchinggenerativemodeling}. The noisy action chunk and flow timestep are first embedded into the backbone's hidden dimension via multi-layer perceptrons (MLPs), then concatenated with the conditioning tokens and processed jointly by the Qwen backbone. An MLP action head predicts the flow-matching velocity at the action positions.

% Let $\mathbf{a} \in \mathbb{R}^{T_a \times d_a}$ denote a clean action chunk of horizon $T_a$ and per-step dimensionality $d_a$, and let $\boldsymbol{\epsilon} \sim \mathcal{N}(\mathbf{0}, \mathbf{I})$ denote Gaussian noise of the same shape. During training, we sample a flow timestep $\tau \in (0, 1)$, form the noisy action $\mathbf{x}_\tau = \tau \boldsymbol{\epsilon} + (1 - \tau) \mathbf{a}$, and train the network to predict the velocity $\mathbf{u}_\tau = \boldsymbol{\epsilon} - \mathbf{a}$:
% \[
% \mathcal{L}_{\text{flow}} = \mathbb{E}_{\mathbf{a}, \boldsymbol{\epsilon}, \tau} \left[ \left\| \mathbf{v}_\theta(\mathbf{x}_\tau, \tau, \mathbf{c}) - \mathbf{u}_\tau \right\|^2 \right],
% \]
% where $\mathbf{v}_\theta$ is the predicted velocity and $\mathbf{c}$ is the conditioning representation from the encoder.

% At inference, we initialize $\mathbf{x}_1 \sim \mathcal{N}(\mathbf{0}, \mathbf{I})$ and integrate from $\tau = 1$ to $\tau = 0$ using $N$ Euler steps of size $\Delta \tau = -1/N$:
% \[
% \mathbf{x}_{\tau + \Delta \tau} = \mathbf{x}_\tau + \Delta \tau \cdot \mathbf{v}_\theta(\mathbf{x}_\tau, \tau, \mathbf{c}).
% \]

\subsection{Data Sources and Representation}
\label{sec:data}
\subsubsection{Egocentric Human Video} Egocentric human video is, by a wide margin, the most abundant source of dexterous manipulation data: it captures the same first-person viewpoint a head-mounted robot camera would observe, spans a vast range of scenes and tasks, and comes with naturally co-occurring hand motion that can be recovered through pose estimation. CAIP trains on 32,041 hours of annotated human egocentric data, collected in both lab ($\sim$1,000 hours, with wrist views) and in-the-wild environments ($\sim$31,000 hours, egocentric view only). We additionally include a small amount of tabletop humanoid manipulation data ($\sim$88 hours) for embodiment diversity and extended wrist-view coverage, which is largely absent from our egocentric sources. This data is collected with a different embodiment and environment than our downstream evaluation setting. The diversity of the data on which we train, in terms of both the environment and task, enables our vision encoder to learn latent representations that can be used downstream to great effect. More details are provided in~\Cref{app:pretraining_data}.

\subsubsection{Action Representation}
\minisection{Hand Pose Representation}
We use end-effector keypoints to represent hand pose, allowing us to form action chunks that are analogous to downstream robotic policy outputs. Specifically, we represent the hand poses at each timestep as a set of 42 keypoints (21 per hand, including the wrist), with each keypoint expressed as an $\mathrm{SE}(3)$ transform, following the MANO hand convention~\citep{Romero_2017}.

These hand poses are collected through various means. The largest portion of our data was annotated using pose estimation techniques, while our in-lab data was collected using the Manus Metagloves Pro and Vive Ultimate Trackers, making these annotations fine-grained and high-quality.

\minisection{Action Chunking}
To convert these poses into actions analogous to what downstream policies must produce, we emulate end-effector delta control. Given a base frame at time $t$, we define the ``action'' at offset $i$ as the relative $\mathrm{SE}(3)$ transform between the pose at time $t$ and the pose at time $t + i$, for $i = 1, \ldots, T$. The full action chunk $A$ is thus a tensor of $T \times 42$ $\mathrm{SE}(3)$ transforms. In practice, we use $T = 64$, thus capturing roughly two seconds of future hand motion at 30\,Hz. 
\newcommand{\mainexptable}{
\begin{table}[t]
\centering
\small
\setlength{\tabcolsep}{4pt}
\setlength{\aboverulesep}{0pt}
\setlength{\belowrulesep}{0pt}
\renewcommand{\arraystretch}{1.15}
\caption{Performance comparison across six real-world manipulation tasks. Each task is evaluated through 12 trials, and results are reported as success rates (\%).}
\begin{tabular}{lccccccc}
\toprule
Method & Fold Shorts & Pour & Pick Fruits & Dispense Soap & Turn On Lamp & Pull Tissue & Avg. \\
\midrule
R3M~\citep{nair2022r3muniversalvisualrepresentation}
& 14.58 & 12.50 & 2.08 & 29.17 & 8.33 & 37.50 & 17.36 \\
Qwen3.5 ViT~\citep{qwen3.5}
& 27.08 & 22.92 & \textbf{60.42} & 72.92 & 8.33 & 12.50 & 34.03 \\
VideoMAE~\citep{tong2022videomaemaskedautoencodersdataefficient}
& 22.92 & 52.08 & 0.00 & 37.50 & 25.00 & 18.75 & 26.04 \\
VC-1~\citep{majumdar2023searchartificialvisualcortex}
& 18.75 & 56.25 & 0.00 & 62.50 & 0.00 & 22.92 & 26.74 \\
MVP~\citep{xiao2022maskedvisualpretrainingmotor}
& 54.17 & 62.50 & 2.08 & 93.75 & 8.33 & 31.25 & 42.01 \\
DINOv2~\citep{oquab2024dinov2learningrobustvisual}
& 22.92 & 81.25 & 52.08 & 50.00 & 25.00 & 20.83 & 42.01 \\
SigLIP~\citep{zhai2023sigmoidlosslanguageimage}
& 12.50 & 70.83 & 37.50 & 83.33 & 25.00 & 25.00 & 42.36 \\
SigLIP 2~\citep{tschannen2025siglip2multilingualvisionlanguage} 
& 4.17 & 35.42 & 52.08 & 93.75 & 50.00 & 25.00 & 43.40 \\
\specialrule{1.5pt}{0pt}{0pt}
\rowcolor{gray!20}
\textbf{CAIP (Ours)}
& \textbf{68.75}
& \textbf{83.33}
& 56.25
& \textbf{100.00}
& \textbf{75.00}
& \textbf{72.92}
& \textbf{76.04} \\
\bottomrule
\end{tabular}
\label{tab:real}
\end{table}
}

\section{Experiments}
\label{sec:exp}
We evaluate CAIP on both downstream policy performance and on representation quality via action retrieval. Additional experiments and analyses are provided in~\Cref{app:additional_exp}.

\mainexptable
\subsection{Real-world Evaluation}
\label{exp:real}

\minisection{Experimental Setup}
We evaluate policies on a real-world setup (see~\Cref{app:hardware_setup}) consisting of a Dexmate Vega bimanual manipulator equipped with two 22-DoF Sharpa Wave dexterous hands. Visual observations are captured from three cameras: the Vega's built-in stereo head camera (ZED X Mini) and two ZED X One S-Wide monocular cameras mounted on the wrists. Actions use end-effector delta control for the arms, and absolute joint control for the fingers. This setup is challenging as the hands are highly dexterous, and we train policies from scratch with only 200 demonstrations per task (150 for pour). Policies are evaluated over 12 trials across six manipulation tasks. Task descriptions, scene configurations, and success criteria are detailed in~\Cref{app:downstream_policy_and_eval}.

\minisection{Baselines}
We compare CAIP against a representative set of vision encoders spanning self-supervised, language-supervised, video, and robotics-pretrained representations: R3M~\citep{nair2022r3muniversalvisualrepresentation}, MVP~\citep{xiao2022maskedvisualpretrainingmotor}, DINOv2~\citep{oquab2024dinov2learningrobustvisual}, SigLIP~\citep{zhai2023sigmoidlosslanguageimage},
SigLIP 2~\citep{tschannen2025siglip2multilingualvisionlanguage},
VideoMAE~\citep{tong2022videomaemaskedautoencodersdataefficient}, VC-1~\citep{majumdar2023searchartificialvisualcortex}, and the native Qwen3.5-0.8B vision encoder~\citep{qwen3.5}. For each baseline, we train a policy from scratch on top of the frozen encoder, as described in~\Cref{method:downstream}. For encoders without a native text tower (R3M, MVP, DINOv2, VideoMAE, VC-1), we use CLIP~\citep{radford2021learningtransferablevisualmodels} to embed the language instruction. All policies share the same downstream architecture, training data, and optimization schedule; only the vision encoder differs across runs. Baseline implementation details are provided in~\Cref{app:baselines}.

We also experimented with direct action regression as a pre-training objective, both from the pooled text-conditioned image embedding via an MLP head and from the full ViT patch sequence via a transformer decoder. Neither variant produced useful representations (details provided in~\Cref{app:baselines}).

\minisection{Results}~\Cref{tab:real} reports per-task success rates across the six manipulation tasks. CAIP achieves the highest average success rate at \textbf{76\%}, outperforming the strongest baseline (SigLIP 2, 43.4\%) by over 30 points. CAIP attains the top success rate on five of the six tasks; in contrast, the baselines do not exhibit consistent behavior: encoders that perform competitively on one task (e.g., MVP at 93.8\% on Dispense Soap, DINOv2 at 81.3\% on Pour) degrade significantly on others. This variance suggests that since these representations are not manipulation-centric, they produce features that may suit some tasks over others. \Ours{}, by contrast, maintains strong performance across all tasks, indicating that our action-centric contrastive pre-training produces representations that generalize. Notably, CAIP outperforms the larger SigLIP 2 SO400M baseline despite being initialized from the smaller SigLIP 2 ViT-L backbone, isolating the gain to our action-centric pre-training. Qualitative rollouts and failure-mode analysis are provided in~\Cref{app:additional_exp}.

\subsection{Zero-Shot Action Classification}
\label{sec:eval:act-classification}
% We evaluate transfer beyond the training distribution using a discrete action-prediction task on a held-out egocentric dataset collected from a source not seen during pre-training. The dataset consists of human manipulation activities and is fully disjoint from all data used to train \Ours{} or the baselines.
To directly evaluate how well \Ours{}'s learned representations generalize beyond the pre-training distribution, we devise an action classification task on a held-out egocentric dataset. This dataset consists of human manipulation activities and is completely disjoint from any training data used.

To construct the task, we cluster held-out action chunks in raw action space using K-means with $K=50$. We then evaluate each frozen vision encoder under two protocols.

\minisection{Linear probing} For each vision encoder (SigLIP, DINOv2, MVP, R3M, and \Ours{}), we train a logistic regression classifier on top of frozen image features to predict the K-means cluster label. We sweep the number of training samples per class from 1 to 256 to measure data efficiency.

\minisection{Zero-shot retrieval (\Ours{} only)} \Ours{} supports retrieval without supervised adaptation. We define each cluster prototype as the mean action embedding over all cluster members, and predict cluster assignments using argmax cosine similarity between the prototypes and image features.

\begin{wrapfigure}{r}{0.45\textwidth}
    \centering
    \vspace{-0.6cm}
    \includegraphics[width=0.45\textwidth]{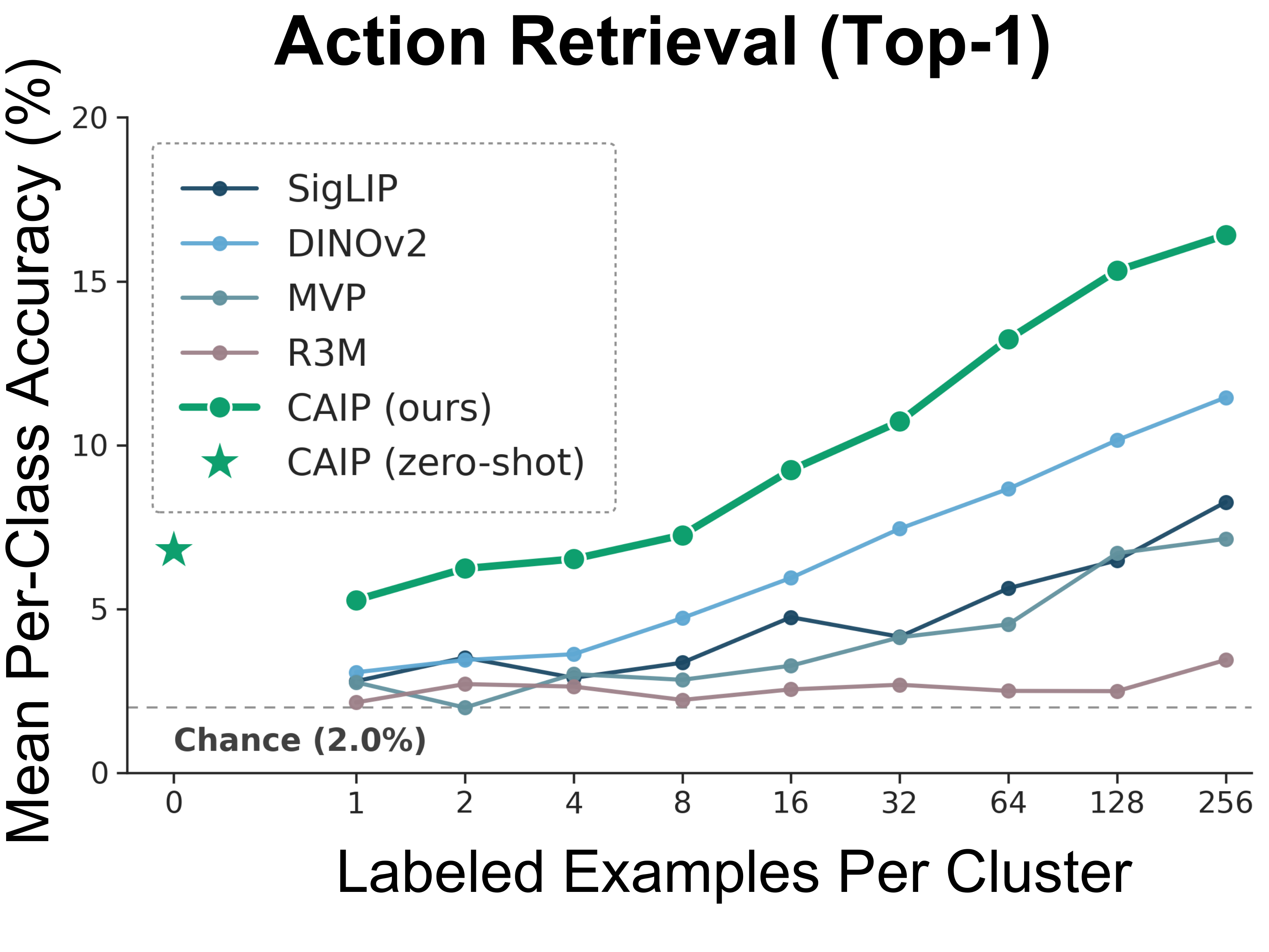}
    \vspace{-0.7cm}
    \caption{Linear probe and zero-shot action classification on the held-out dataset.}
    \label{fig:linear_probe}
    \vspace{-2em}
\end{wrapfigure}
As shown in Figure~\ref{fig:linear_probe}, \Ours{} consistently outperforms all baselines, including the strongest, DINOv2, by a substantial margin across the full data-efficiency curve. Notably, zero-shot retrieval with \Ours{}, despite requiring no in-domain supervision, exceeds the linear-probe performance of every baseline up to 16 samples per class. These results suggest that \Ours{}'s learned representations transfer effectively to unseen data and capture semantically meaningful structure for action understanding.

\subsection{Environmental Robustness Analysis}
\label{exp:subsec:env-changes}
We evaluate the robustness of our policies under environmental perturbations. Specifically, we vary the scene lighting and add distractor objects, then evaluate each policy in the perturbed environment (see~\Cref{app:downstream_policy_and_eval:env-robustness}). Because all policies are trained only on demonstrations collected under standard conditions, any change in success rate reflects the sensitivity of the learned visual representation to these perturbations.

\minisection{Lighting Variation}
We consider two lighting perturbations. In the \emph{Light} condition, we add an extra bulb above the scene in front of the robot, which casts shadows across the workspace. In the \emph{Dark} condition, we reduce the intensity of the standard scene lighting. 

\minisection{Distractors}
We add two distractor objects to the scene: a red book and a multi-colored Hanoi toy tower. Both are placed within the manipulation area so that they remain clearly visible in the egocentric camera, and their locations are randomized across trials.

\minisection{Results}
As shown in Tables~\ref{tab:lighting-robustness} and~\ref{tab:distractor-robustness}, CAIP achieves the highest success rate under every perturbation. While all encoders degrade relative to the original setting, CAIP remains the strongest, outperforming Qwen3.5 ViT by roughly 20--30\% and degrading far less than MVP. We note that Qwen3.5 ViT retains a larger \emph{relative} fraction of its original performance under some perturbations, but its original success rate is also much lower, leaving less room to degrade. In absolute terms, CAIP is the most robust of the encoders we compare.

\begin{table}[t]
\centering
\tablestyle{0.1pt}{1.1}
\caption{Policy success rates (\%) under varying lighting conditions. All policies are trained only on demonstrations collected under original lighting. Average columns report the mean success rate across all three tasks under each lighting condition. Best result per column is shown in \textbf{bold}.}
\label{tab:lighting-robustness}
\small
\setlength{\tabcolsep}{4pt}
\begin{tabular}{l ccc c ccc c ccc c ccc}
\toprule
& \multicolumn{3}{c}{Turn on Lamp} & & \multicolumn{3}{c}{Dispense Soap} & & \multicolumn{3}{c}{Fold Shorts} & & \multicolumn{3}{c}{Average} \\
\cmidrule(lr){2-4} \cmidrule(lr){6-8} \cmidrule(lr){10-12} \cmidrule(lr){14-16}
Encoder & Orig. & Light & Dark & & Orig. & Light & Dark & & Orig. & Light & Dark & & Orig. & Light & Dark \\
\midrule
Qwen3.5 ViT        & 8.33 & 0.00 & 16.67 & & 72.92 & 58.33 & 56.25 & & 27.08 & 8.33 & 0.00 & & 36.11 & 22.22 & 24.31 \\
MVP         & 8.33 & 0.00 & 8.33 & & 93.75 & 8.33 & 39.58 & & 54.17 & 8.33 & 4.17 & & 52.08 & 5.56 & 17.36 \\
\specialrule{1.5pt}{0pt}{0pt}
\rowcolor{gray!20}
\textbf{CAIP (ours)} & \textbf{75.00} & \textbf{25.00} & \textbf{33.33} & & \textbf{100.00} & \textbf{93.75} & \textbf{75.00} & & \textbf{68.75} & \textbf{35.42} & \textbf{20.83} & & \textbf{81.25} & \textbf{51.39} & \textbf{43.06} \\
\end{tabular}
\end{table}

\begin{table}[t]
\centering
\tablestyle{0.1pt}{1.1}
\caption{Policy success rates (\%) under the influence of distractors. All policies are trained only on demonstrations collected without distractors. Average columns report the mean success rate across all three tasks under each distractor condition. Best result per column is shown in \textbf{bold}.}
\label{tab:distractor-robustness}
\small
\setlength{\tabcolsep}{4pt}
\begin{tabular}{l cc c cc c cc c cc}
\toprule
& \multicolumn{2}{c}{Turn on Lamp} & & \multicolumn{2}{c}{Dispense Soap} & & \multicolumn{2}{c}{Fold Shorts} & & \multicolumn{2}{c}{Average} \\
\cmidrule(lr){2-3} \cmidrule(lr){5-6} \cmidrule(lr){8-9} \cmidrule(lr){11-12}
Encoder & Orig. & 2 Distractors & & Orig. & 2 Distractors & & Orig. & 2 Distractors & & Orig. & 2 Distractors \\
\midrule
Qwen3.5 ViT        & 8.33 & 8.33 & & 72.92 & 60.42 & & 27.08 & 16.67 & & 36.11 & 28.47 \\
MVP         & 8.33 & 0.00 & & 93.75 & 12.50 & & 54.17 & 16.67 & & 52.08 & 9.72 \\
\specialrule{1.5pt}{0pt}{0pt}
\rowcolor{gray!20}
\textbf{CAIP (ours)} & \textbf{75.00} & \textbf{33.00} & & \textbf{100.00} & \textbf{85.42} & & \textbf{68.75} & \textbf{39.58} & & \textbf{81.25} & \textbf{52.78} \\
\end{tabular}
\end{table}

\subsection{Scaling Ablations}

We study the effect of scaling the vision encoder across ViT-B, ViT-L, and ViT-SO400M on three manipulation tasks. As shown in~\Cref{app:vision_encoder_scaling}, increasing encoder scale leads to substantial performance improvements, with the transition from ViT-B to ViT-L yielding the largest gain of over 30\% on average. Based on this ablation, we select ViT-L as the primary vision encoder for all experiments, as it provides the best trade-off between performance, model size, and inference speed. Additional scaling ablations and analyses are provided in~\Cref{app:additional_exp}.

\section{Related Work}

\minisection{Internet-Scale Image-Language Encoders} The evolution of visual representations for robotic manipulation has progressed from convolutional neural networks (CNNs) trained directly on raw pixels~\citep{levine2016endtoendtrainingdeepvisuomotor, finn2016deepspatialautoencodersvisuomotor,rahmatizadeh2018visionbasedmultitaskmanipulationinexpensive} to transformer-based architectures with substantially greater representational capacity and scalability~\citep{dosovitskiy2021imageworth16x16words, brohan2023rt1roboticstransformerrealworld}. Yet despite these architectural advances, the ViT backbones powering modern VLAs~\citep{rt22023arxiv, black2026pi0visionlanguageactionflowmodel, intelligence2025pi05visionlanguageactionmodelopenworld} are pre-trained on internet-scale image-language data that is fundamentally misaligned with the demands of manipulation.

% CLIP~\citep{radford2021learningtransferablevisualmodels} established the contrastive image-text recipe that inspires our own architecture, but its training distribution consists almost entirely of web images and their captions, with no physical interaction. SigLIP~\citep{zhai2023sigmoidlosslanguageimage} refines the objective with a pairwise sigmoid loss but is trained on WebLI~\citep{wang2025scalingpretrainingbilliondata}, which also lacks physical interaction data. As a result, these vision encoders learn the semantic meaning of various scenes, but not what real-world manipulation in these scenes entails.

CLIP~\citep{radford2021learningtransferablevisualmodels} introduced the contrastive image-text training paradigm that inspires our architecture, but is trained primarily on images and captions without physical interaction. SigLIP~\citep{zhai2023sigmoidlosslanguageimage} improves the objective with a pairwise sigmoid loss, but is likewise trained on WebLI~\citep{wang2025scalingpretrainingbilliondata}, which lacks interaction. Thus, these encoders capture semantics, but not the actions required for real-world manipulation.

Since the vast majority of modern vision encoders are pre-trained on 
internet-scale image-language corpora, their limitations propagate 
directly into state-of-the-art VLAs. Systems such as RT-2~\citep{rt22023arxiv}, 
$\pi_{0.5}$~\citep{intelligence2025pi05visionlanguageactionmodelopenworld}, OpenVLA~\citep{kim24openvla}, and 
GR00T~\citep{nvidia2025gr00tn1openfoundation} all build upon large-scale vision or vision-language backbones~\citep{zhai2023sigmoidlosslanguageimage, oquab2024dinov2learningrobustvisual,beyer2024paligemmaversatile3bvlm, li2025eagle2buildingposttraining} originally optimized for semantic internet understanding rather than manipulation-centric perception.

\minisection{Egocentric Representation Learning}
More recent efforts have sought to mitigate data mismatch by pre-training specifically on manipulation settings. Egocentric human video is the most abundant such data source available, with corpora like Ego4D~\citep{grauman2022ego4daroundworld3000}, Epic-Kitchens~\citep{damen2020rescalingegocentricvisioncollection}, and EgoDex~\citep{hoque2025egodexlearningdexterousmanipulation} offering thousands of hours of hand-object interaction. The rise of such datasets has motivated a line of work building visual representations directly on top of egocentric manipulation data, with many of these encoders showing substantive gains over backbones pre-trained on image-text pairs. 
% We see this progress as compelling evidence that the input distribution matters, but believe there is further potential left on the table: while the data has shifted toward manipulation, the learning objectives have not. Inheriting action-agnostic losses from internet-scale pretraining yields encoders that capture what manipulation looks like in visual space without modeling what manipulation is in action space.

R3M~\citep{nair2022r3muniversalvisualrepresentation} was among the first frozen backbones pre-trained explicitly on human video data, using time-contrastive learning and video-language alignment over 3,700 hours of Ego4D. Its time-contrastive loss encourages the encoder to produce embeddings such that temporally-close frames are similar, and temporally-distant frames are more dissimilar. This self-supervised approach allows the encoder to capture intra-frame relationships, but it does not expose the encoder to ground-truth action or poses during pre-training.
MVP~\citep{xiao2022maskedvisualpretrainingmotor, radosavovic2022realworldrobotlearningmasked} and VC-1~\citep{majumdar2023searchartificialvisualcortex} take a more direct route by porting masked autoencoding~\citep{he2021maskedautoencodersscalablevision} onto egocentric and manipulation-relevant images. The data is appropriate, but the objective is identical to its internet-scale counterpart: predict missing pixels, regardless of whether those pixels matter for robotic control purposes. HRP~\citep{srirama2024hrphumanaffordancesrobotic} is a step toward explicit pre-training with action annotations, fine-tuning encoders to predict hand-object affordances, like contact points, future hand poses, and active objects. However, these supervision signals are still distant from the downstream actions policies must produce.

\minisection{Action-Aware Visual Representation Learning} A growing body of work learns latent action representations directly from video, without ground-truth robot action labels.
One family of methods discretizes inter-frame transitions into latent action tokens via VQ-VAE objectives, then pre-trains a VLA to predict them from observations and language \citep{ye2025latentactionpretrainingvideos, chen2025motolatentmotiontoken, dai2026conlacontrastivelatentaction}.
A related line, such as UniVLA~\citep{bu2025univlalearningacttaskcentric} and CLAP~\citep{zhang2026clapcontrastivelatentaction}, incorporates contrastive objectives or language grounding to suppress task-irrelevant visual dynamics and improve cross-embodiment transfer. While demonstrating the value of learning from video, these methods rely on ungrounded latent action tokens rather than explicit action supervision, and are designed as full vision-language-action systems, making them difficult to deploy as general-purpose visual backbones for arbitrary downstream policies.

A parallel line of work supervises representation learning with action correspondences on robot data, using contrastive objectives that align visual observations with proprioceptive state-action dynamics \citep{jiang2024robotspretrainrobotsmanipulationcentric, lee2025classcontrastivelearningaction, kim2025contrastiverepresentationregularizationvisionlanguageaction}, or incorporating 3D geometric structure from depth and point cloud observations \citep{liu2026clampcontrastivelearning3d, wang2024visualroboticmanipulationdepthaware}. These methods are constrained by the scale and diversity of robot demonstration data. In contrast, \Ours{} leverages large-scale egocentric human video as its primary pre-training source, thus scaling its visual pre-training beyond what robot datasets alone can support.

% \vspace{-6pt}
\section{Conclusion}
In this work, we show that vision encoders can learn action-centric representations for dexterous manipulation from large-scale egocentric human video, using hand poses as a proxy for end-effector actions. By aligning visual observations with action chunks through a contrastive objective, our method learns representations that are more effective for downstream robotic control than standard semantic or self-supervised pre-training. Our approach combines a ViT image encoder with an action transformer and attention pooling, and is trained on 32,129 hours of egocentric video. The resulting encoder achieves a 76\% average success rate across downstream dexterous manipulation tasks, outperforming strong vision baselines including DINOv2, SigLIP, MVP, and Qwen3.5 ViT. We further show through action retrieval experiments that action-grounded pre-training transfers effectively to unseen data. Overall, our results suggest that large-scale human video paired with action supervision provides a scalable path toward manipulation-oriented visual pre-training.

\section{Limitations and Future Work} 
\label{sec:limitations}

% While CAIP yields strong empirical gains, several limitations remain that point toward promising directions for future work.

\minisection{Negative sampling under continuous action structure} 
% Our contrastive objective treats all off-diagonal image--action pairs within a batch as negatives, regardless of how similar the underlying actions actually are. In practice, two trajectories drawn from different timesteps or scenes may produce nearly identical hand motions (e.g., two pouring actions, or two reaches toward similar objects), yet our loss pushes their representations apart. This weakens the learning signal and likely caps the quality of the learned representation, especially as the action space is continuous and densely populated. Future work could explore soft contrastive objectives that weight negatives by their action-space distance to the anchor, such as Similarity Contrastive Estimation~\cite{denize2022similaritycontrastiveestimationselfsupervised}, which replaces the binary positive/negative label with a continuous similarity distribution over the batch. \yuvan{how is this?}
Our contrastive objective treats all off-diagonal image--action pairs within a batch as negatives, regardless of their physical similarity. In practice, distinct trajectories drawn from different timesteps or scenes may feature similar hand motions (e.g., two pouring actions or two reaches toward similar targets), yet the loss will actively push their representations apart. This assumption can weaken the learning signal and constrain representation quality. Future work could explore soft contrastive objectives that weight negatives by their action-space distance to the anchor.

%, which replaces binary positive--negative labels with a continuous similarity distribution over the batch.

% \minisection{Anthropomorphic bias from the hand-pose proxy} Our action representation is built around the 42-keypoint MANO skeleton, which biases the learned representation toward anthropomorphic morphology. While this aligns well with five-fingered dexterous end-effectors like the Sharpa hands used in our evaluation, it is unclear how well CAIP would transfer to embodiments that deviate substantially from human hand kinematics, such as three-fingered hands, parallel-jaw grippers, or other non-anthropomorphic dexterous designs. Future work should evaluate CAIP across a broader range of end-effector morphologies to characterize the regime in which human hand pose serves as a useful action proxy.

\minisection{Anthropomorphic bias from the hand-pose proxy} Our action representation is built around the 42-keypoint MANO skeleton, which biases the learned features toward human hands. While this aligns well with five-fingered end-effectors like the Sharpa hands, CAIP's transferability to embodiments such as parallel-jaw grippers or three-fingered claws remains an open question. Future work should evaluate CAIP across a broader range of end-effector morphologies to characterize the regime in which human hand pose serves as a useful action proxy.

% \minisection{Frozen-encoder evaluation} All downstream policies in this work use the CAIP encoder as a frozen feature extractor. This isolates the contribution of pre-training cleanly, but leaves open the question of whether end-to-end fine-tuning would amplify or erode CAIP's advantage over baselines. It is possible that other encoders close the gap under fine-tuning, or conversely, that CAIP's action-aligned features serve as a particularly strong initialization. We leave a systematic ablation of frozen versus fine-tuned encoders to future work.

% \minisection{Frozen-encoder evaluation} We evaluate downstream policies using CAIP exclusively as a frozen feature extractor. This strategy cleanly isolates the pre-training contribution, but whether end-to-end fine-tuning would enhance or diminish CAIP's performance advantage remains an open question. Fine-tuning could allow baselines to close the gap or demonstrate that action-aligned features provide a superior initialization. We leave a systematic ablation of fine-tuning regimes to future work.
%===============================================================================

\acknowledgments{We thank Sharpa and Dexmate for their continued technical support, including equipment maintenance and software updates. UC Berkeley authors were supported in part by the Berkeley Artificial Intelligence Research Humanoid Intelligence Center (BAIR HIC). Sapienza University acknowledges funding from Panasonic and from the Sapienza grant RG123188B3EF6A80 (CENTS).}

\clearpage
% The acknowledgments are automatically included only in the final and preprint versions of the paper.
% \acknowledgments{If a paper is accepted, the final camera-ready version will (and probably should) include acknowledgments. All acknowledgments go at the end of the paper, including thanks to reviewers who gave useful comments, to colleagues who contributed to the ideas, and to funding agencies and corporate sponsors that provided financial support.}

%===============================================================================

\bibliography{references}  % .bib
\clearpage
\appendix
\section*{Appendix}
\section{Additional Experiments}
% scaling etc.
% failure case analysis
\label{app:additional_exp}

\subsection{Saliency Visualization Details}
\label{app:saliency}

\begin{figure}
    \centering
    \includegraphics[width=0.835\linewidth]{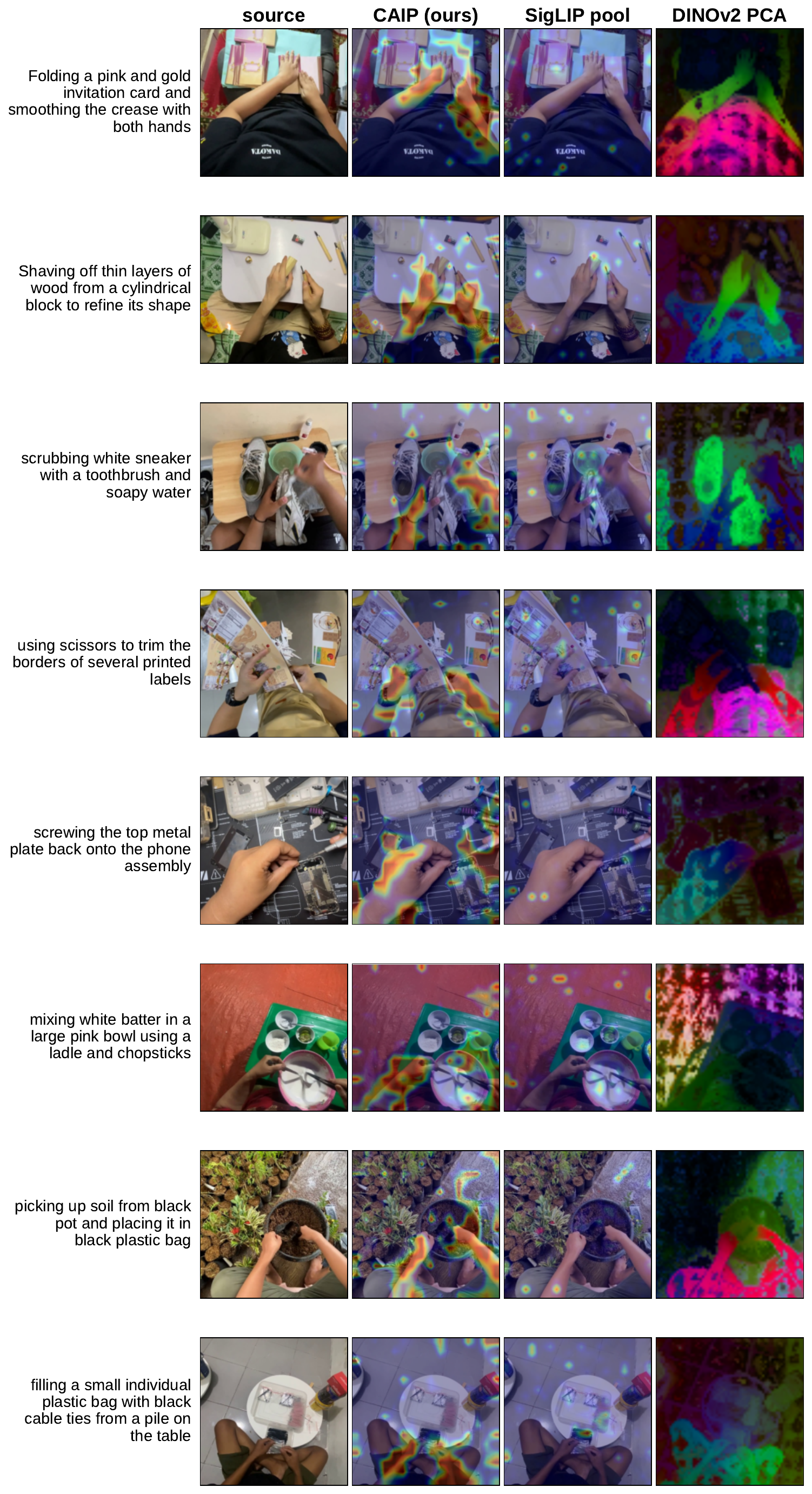}
    \caption{\textbf{Saliency across vision encoders} on held-out egocentric
manipulation frames. Columns: input, CAIP (ours), SigLIP, DINOv2. CAIP's
text-conditioned cross-attention pool (aggregated over instruction tokens) focuses
on the hands and manipulated object; SigLIP's text-agnostic learned probe
scatters across background sink patches; DINOv2's per-image PCA segments by
appearance but is instruction-unaware (colors not comparable across rows). See
Appendix~\ref{app:saliency}.}
    \label{fig:appendix-saliency}
\end{figure}

The visualizations in Figure~\ref{fig:teaser} (left) and the per-encoder comparison in Figure~\ref{fig:appendix-saliency} are computed using each encoder's native query mechanism, since the three encoders expose spatial information in fundamentally different ways. Unless otherwise noted, all maps are overlaid on a common square center crop of the input image to ensure spatial alignment across encoders.

\minisection{SigLIP} SigLIP's vision tower terminates in a multi-head attention-pooling (MAP) head whose query is a single learned probe. Consequently, the resulting attention map is text-agnostic. We obtain the probe-to-patch attention weights, reshape them to the patch grid, min--max normalize each head independently, and visualize the mean across heads. The resulting map is overlaid on the image with opacity $\alpha = 0.55$.

\minisection{DINOv2} DINOv2 does not contain an explicit pooling query. Following the standard visualization protocol~\citep{oquab2024dinov2learningrobustvisual}, we extract the final-layer patch tokens and fit PCA independently for each image. The first three principal components are mapped to RGB channels after percentile normalization to expose semantic structure. The first principal component is additionally used as a soft foreground mask that gradually fades the background to black. Since PCA is fit independently for each image, we note that colors are not comparable across examples.

\minisection{CAIP (Ours)} Our encoder architecture ends with a text-conditioned cross-attention pooling layer in which text tokens act as queries and image patches serve as keys and values. The resulting attention is therefore natively instruction-conditioned. We extract the pooling attention tensor 
$$
A \in \mathbb{R}^{H \times L \times N}$$
where $H$ denotes attention heads, $L$ text tokens, and $N$ image patches.

For each non-special text token (excluding padding, BOS, and EOS tokens), we normalize each head by its spatial maximum and then take the maximum across heads, yielding one spatial map per token. We subsequently aggregate across all content tokens using a per-location maximum operation. In practice, this token-aggregate map is visually indistinguishable from the map produced by any individual content token, and therefore provides a concise summary of the encoder's behavior without requiring token selection. For visualization, the aggregated map is percentile clipped (lower/upper percentiles $50/98$) and gamma corrected ($\gamma = 1.6$) before being overlaid on the image.

\minisection{Analysis} The following observations emerge from these visualizations.

\textbf{(i) CAIP's text-conditioned attention focuses on a stable task region.}
Although each text token produces its own attention distribution, the resulting maps are nearly identical across content tokens. Attention consistently concentrates on the hands and the actively manipulated object rather than localizing the specific noun or verb associated with a given token. We attribute this to how the pooling layer is supervised: the contrastive
objective constrains only the pooled output, not the individual per-token
attention rows. Without any token-specific supervision, the per-token maps are
nearly identical to one another, which is the motivation for why we
visualize the aggregate.

\textbf{(ii) Only CAIP consistently highlights task-relevant regions.}
SigLIP's learned probe is text-agnostic and often concentrates on a small number of high-norm patches, frequently in low-information background regions. This behavior is consistent with the register ``sink'' token phenomenon observed in ViTs~\citep{darcet2024visiontransformersneedregisters} and generally does not track the manipulation being performed. DINOv2, by contrast, produces clean semantic segmentations of the scene but, as an unsupervised and instruction-unaware representation, does not reliably identify the task-relevant region.

\subsection{Scaling Vision Encoder Capacity}
\label{app:vision_encoder_scaling}

We study how vision encoder capacity affects downstream policy performance by pre-training CAIP at three scales—ViT-B/16, ViT-L/16, and ViT-SO400M/16—under identical data, training, and optimization settings, and evaluating each on three downstream tasks. Table~\ref{tab:model-scale} reports the results.

Performance improves consistently with encoder scale. The largest gain comes from ViT-B to ViT-L, where average success rate rises from 47.9\% to 81.3\%, an absolute improvement of over 30 points. This jump is driven primarily by the two harder tasks: Turn on Lamp improves from 16.7\% to 75.0\% and Fold Shorts from 54.2\% to 68.8\%, while Dispense Soap, already strong at the smallest scale, saturates at 100\%. The gains suggest that the smaller ViT-B backbone lacks the capacity to learn the fine-grained, action-relevant features required for precise manipulation, whereas ViT-L provides sufficient capacity to capture them.
\begin{table}[t]
\centering
\caption{Policy success rates (\%) using different vision encoder scales. All encoders are pre-trained and evaluated under identical settings.}
\label{tab:model-scale}
\small
\setlength{\tabcolsep}{6pt}
\renewcommand{\arraystretch}{1.15}
\begin{tabular}{lccc}
\toprule
Task & ViT-B/16 & ViT-L/16 & ViT-SO400M/16 \\
\midrule
Turn On Lamp  & 16.67 & 75.00 & \textbf{83.33} \\
Fold Shorts   & 54.17 & 68.75 & \textbf{79.17} \\
Dispense Soap & 72.92 & \textbf{100.00} & \textbf{100.00} \\
\midrule
Average       & 47.92 & 81.25 & \textbf{87.50} \\
\bottomrule
\end{tabular}
\end{table}

Scaling further to ViT-SO400M yields a smaller additional improvement, raising the average from 81.3\% to 87.5\%. Given that ViT-SO400M is substantially larger and slower at both training and inference time, we adopt ViT-L as our main encoder, as it captures most of the benefit of scaling while remaining efficient enough for practical downstream use.

\subsection{Simulation Results}
\label{app:sim_results}

To test whether CAIP's learned representations transfer beyond the embodiment and domain they were trained for, we evaluate on a simulated manipulation benchmark that differs substantially from all of our other experiments. Whereas CAIP is pre-trained on egocentric \emph{dexterous} human manipulation and evaluated in the real world on a bimanual humanoid platform, here we use the ManiSkill2~\citep{gu2023maniskill2} Franka setup: a \emph{single} 7-DoF Franka arm with a parallel-jaw gripper performing tabletop tasks. This represents a large shift in embodiment (single arm vs.\ bimanual dexterous hands), action space, and visual domain (simulation vs.\ real-world egocentric video). We keep the same evaluation protocol as our real-world experiments: the vision encoder is frozen, a policy is trained from scratch on 200 demonstrations per task using the overhead and wrist views, and each (encoder, task) pair is evaluated over 12 trials.

Table~\ref{tab:sim-results} reports per-task success rates. Despite the substantial domain gap, CAIP achieves the highest average success rate (77.8\%), outperforming all baselines including the strongest, SigLIP 2 (72.2\%) and DINOv2 (69.4\%). These results indicate that CAIP's action-centric representations provide a useful prior for manipulation even under a large shift in embodiment and visual domain from the dexterous bimanual setting it was trained on.

\begin{table}[t]
\centering
\caption{Policy success rates (\%) on the ManiSkill2 Franka simulation tasks, a single-arm tabletop setup that differs substantially from CAIP's egocentric dexterous pre-training domain. Each (encoder, task) pair is evaluated over 12 trials. Best result per column in \textbf{bold}.}
\label{tab:sim-results}
\small
\setlength{\tabcolsep}{6pt}
\renewcommand{\arraystretch}{1.15}
\begin{tabular}{lcccc}
\toprule
Encoder & Lift Peg & Stack Cubes & Push Cube & Average \\
\midrule
VC-1      & 16.67 & 0.00  & 91.67 & 36.11 \\
R3M       & 25.00 & 16.67 & 91.67 & 44.44 \\
MVP       & 33.33 & 0.00  & \textbf{100.00} & 44.44 \\
DINOv2    & 58.33 & 50.00 & \textbf{100.00} & 69.44 \\
SigLIP 2  & 58.33 & \textbf{58.33} & \textbf{100.00} & 72.22 \\
\specialrule{1.5pt}{0pt}{0pt}
\rowcolor{gray!20}
\textbf{CAIP (ours)} & \textbf{75.00} & \textbf{58.33} & \textbf{100.00} & \textbf{77.78} \\
\bottomrule
\end{tabular}
\end{table}

\subsection{Scaling Vision Encoder Data}
\label{app:vision_encoder_data_scaling}

We next study how the amount of pre-training data affects downstream performance, holding the encoder fixed at ViT-L/16 and varying the fraction of pre-training data used. We pre-train CAIP on 20\%, 50\%, and 100\% of the full dataset, training each for the same number of epochs so that the comparison reflects the amount of unique data seen by the model. Each encoder is evaluated on the ManiSkill2 Franka tasks under the same protocol as Appendix~\ref{app:sim_results} (frozen encoder, policy trained from scratch on 200 demonstrations per task, 12 trials per task). Table~\ref{tab:data-scale} reports the results.

Performance improves monotonically with pre-training data. Average success rate rises from 50.0\% at 20\% of the data to 61.1\% at 50\%, and to 77.8\% at full scale. The gains are consistent across all three tasks: every task improves at each step, with the hardest task, Stack Cubes, more than doubling from 25.0\% to 58.3\%. Notably, we observe no sign of saturation, with performance still climbing at the full data scale. This result suggests that CAIP would continue to benefit from additional pre-training data beyond what we currently use. This supports our central premise: that large-scale egocentric human video is a valuable and scalable source of supervision for action-centric visual representations.

\begin{table}[t]
\centering
\caption{Policy success rates (\%) as a function of pre-training data scale, holding the encoder fixed at ViT-L/16. All runs are trained for the same number of epochs. Evaluated on the ManiSkill2 Franka tasks over 12 trials each. Best result per column in \textbf{bold}.}
\label{tab:data-scale}
\small
\setlength{\tabcolsep}{6pt}
\renewcommand{\arraystretch}{1.15}
\begin{tabular}{lcccc}
\toprule
Pre-training data & Lift Peg & Stack Cubes & Push Cube & Average \\
\midrule
20\%  & 58.33 & 25.00 & 66.67  & 50.00 \\
50\%  & 66.67 & 33.33 & 83.33  & 61.11 \\
100\% & \textbf{75.00} & \textbf{58.33} & \textbf{100.00} & \textbf{77.78} \\
\bottomrule
\end{tabular}
\end{table}

\subsection{Failure Case Analysis}
We observe a few recurring failure modes in policies trained with the \Ours{} encoder that highlight potential shortcomings. In the pouring task, a common failure was to grasp successfully but then begin pouring before the two cups were aligned. We hypothesize that this stems from both cups being the same color (red): when the cups are close together but not yet aligned, the visual features may closely resemble those of the aligned configuration. In the fold-shorts task, the policy would sometimes fail to attempt the second fold, likely because the visual features after the first fold resemble those of the completed state. We observed both failure modes in policies trained with other vision encoders as well, suggesting they reflect general perceptual ambiguities of the tasks.

\section{Hardware Setup}
\label{app:hardware_setup}
\begin{figure}[h]
    \centering
    \includegraphics[width=\textwidth]{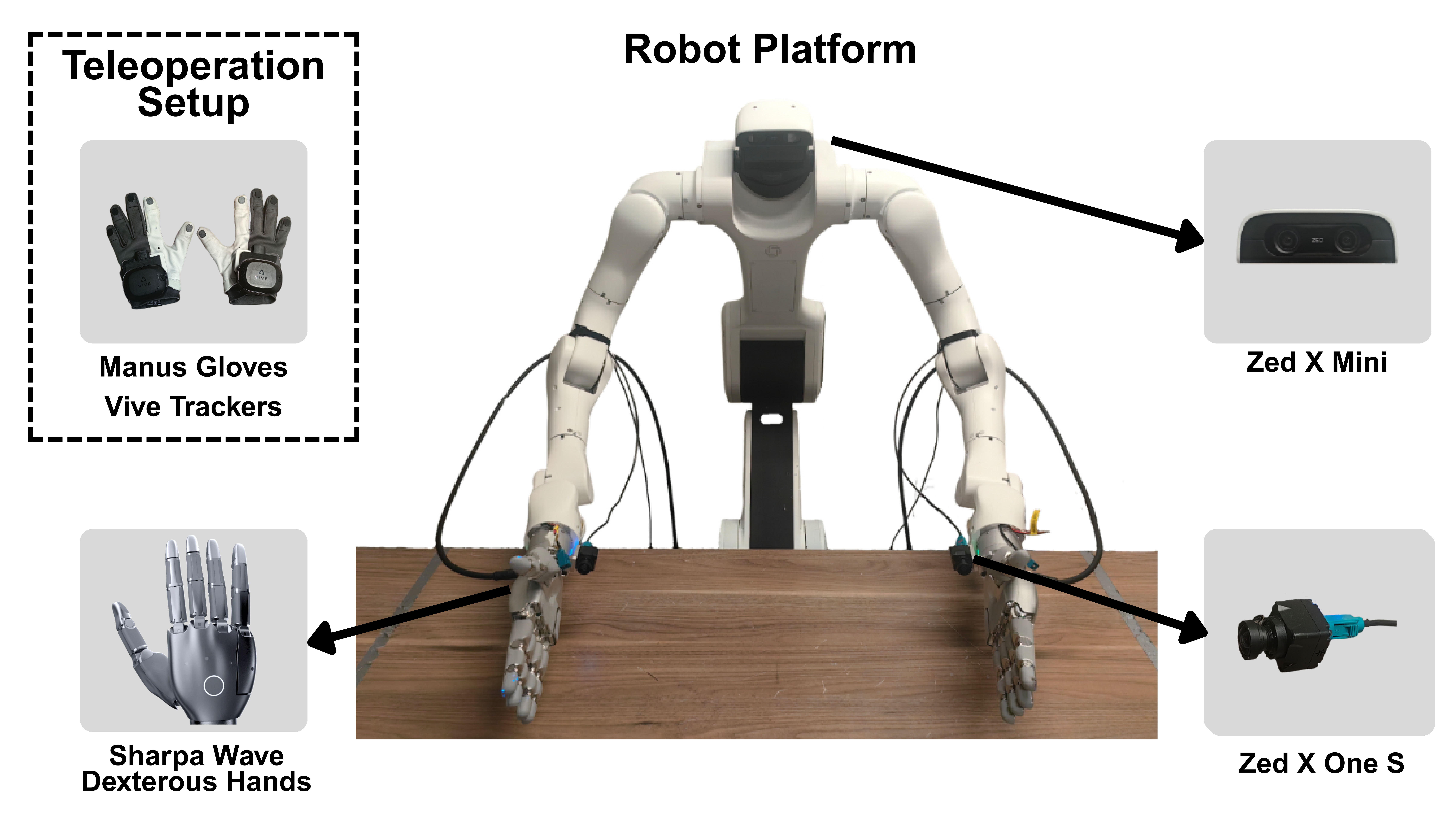}
    \caption{CAIP Hardware Setup.}
    \label{fig:caip_hardware}
\end{figure}

\subsection{Embodiment}
Our physical setup is a fixed-base bimanual manipulator built on the Dexmate Vega platform. We mount each arm with the Sharpa Wave dexterous hands to allow for precise manipulation (see~\Cref{fig:caip_hardware}).

\subsubsection{Dexmate Vega}
The Dexmate Vega is a dual-arm mobile manipulation platform with 36 total degrees of freedom (DoF), spanning an omnidirectional wheeled base, a foldable torso, an articulated head, and two 7-DoF arms. For all experiments in this work, we operate the Vega as a fixed-base bimanual manipulator: the base, torso, and head joints are held static, and we drive only the two 7-DoF arms (14 joints in total). We replace the platform's native end-effectors with Sharpa Wave hands, so each 7-DoF arm provides full $SE(3)$ positioning of its attached dexterous hand. Arm motion is commanded as relative end-effector pose targets (Section~\ref{app:hardware_setup:control}).

\subsubsection{Sharpa Wave}
Each arm is equipped with a Sharpa Wave hand, an anthropomorphic five-fingered end-effector with 22 active degrees of freedom built at 1:1 human size and scale. The hand uses a tendon-driven transmission and is fully actuated, allowing all finger joints to be commanded directly in joint space without the mechanical coupling or masked joints common to lower-DoF designs. We control the 22 joints via absolute position targets. Although the Sharpa Wave integrates onboard fingertip tactile sensing, we do not use any tactile or proprioceptive signals in this work.

\subsection{Camera Setup}
\label{app:hardware_setup:cameras}
Visual observations are captured from three cameras. A ZED X Mini stereo camera mounted on the Vega head provides an egocentric view that approximates the first-person viewpoint of our egocentric pre-training data, and two ZED X One S monocular cameras (wide-view variant) are mounted on the wrists to capture close-range views of each hand that are otherwise occluded from the head. We position the head camera to cover the full reachable workspace in front of the robot, and the wrist cameras to keep the fingers in clear view with minimal palm occlusion. All three streams are RGB and captured at $640 \times 360$ resolution; we do not use depth.

\subsection{Teleoperation}
\label{app:hardware_setup:teleop}
We collect demonstrations through a teleoperation interface that maps the operator's wrist and finger motion onto the bimanual platform. The interface shares the same control pipeline as policy rollout, ensuring that demonstrated and executed policy actions occupy a consistent action space.

\subsubsection{Pose Tracking}
The operator wears a pair of Manus gloves, each augmented with a Vive Ultimate Tracker attached via a custom 3D-printed mount that rigidly fixes the tracker to the back of the glove. The two devices supply complementary signals. Each Vive tracker reports the 6-DoF $SE(3)$ pose of the corresponding wrist in the Vive world frame, which drives arm control. The Manus glove reports 3D finger keypoints expressed relative to the wrist (hand-base) frame, which drive hand control.

We map tracked wrist motion to the robot using relative end-effector (delta) control. Given consecutive wrist poses in the Vive world frame, we compute the relative $SE(3)$ transform and apply it as a delta to the current robot end-effector pose to obtain a target pose.
\subsubsection{Control}
\label{app:hardware_setup:control}
\textbf{Arms.} Given a target end-effector pose (either from teleoperation or produced by a policy), we compute the corresponding 7-DoF arm joint angles using differential inverse kinematics implemented in Pink~\citep{pink}, which builds on the Pinocchio rigid-body dynamics library~\citep{pinocchio}. The resulting joint commands are smoothed with a low-pass filter and passed to the manufacturer's low-level cascade PID controller. A high-level loop generates target poses at $30$~Hz---from the teleoperator during data collection, or from the policy at inference---and asynchronously updates the targets tracked by a $300$~Hz low-level control thread.

\textbf{Hands.} We retarget the Manus 3D finger keypoints, expressed in the wrist frame, onto the Sharpa Wave joint space using a manufacturer-provided differential inverse kinematics package built on Pinocchio~\citep{pinocchio} and CasADi~\citep{casadi}, yielding absolute joint-position targets for the hand's 22 actuated DoF. This establishes a direct correspondence between the operator's finger configuration and the commanded hand pose.

\section{Pre-training Details}

\subsection{Data}
\label{app:pretraining_data}
% - Per-dataset description in one combined paragraph or compact table:
%   EgoDex, in lab human/robot, in the wild egocentric (mecka) 
%   Include hours/frames per source, total
% - Upsampling factors used and resulting effective composition
% - Action representation: T × 42 × 9, delta-base, per-dim normalization
% - Preprocessing: image resolution, augmentations, text format
\minisection{Pre-training Data} We pre-train on a mixture of four egocentric manipulation datasets spanning both in-lab demonstrations and large-scale in-the-wild videos (Table~\ref{tab:pretrain_data}). The in-lab robot demonstrations were collected on the Galaxea R1Pro humanoid robot with 22-DoF Sharpa dexterous hands. All sources are converted to a unified hand-action representation. EgoDex~\citep{egodex} and the large-scale in-the-wild dataset are recorded at 30\,Hz, while the in-lab datasets are recorded at 20\,Hz. In total, the corpus contains approximately \textbf{3.46\,B frames} ($\sim$32{,}000 hours) of egocentric interaction data. The in-the-wild dataset contributes the majority of the raw volume ($96.6\%$ of all frames). To prevent this source from dominating training, we apply dataset-specific upsampling factors, yielding an effective sampling distribution of $10\%$ EgoDex, $3\%$ in-lab human, $2\%$ in-lab robot, and $85\%$ large-scale in-the-wild data. These in-lab sources also carry higher-fidelity supervision: EgoDex hand poses are tracked with the Apple Vision Pro headset, the in-lab human poses with Manus Metagloves Pro gloves and Vive trackers, and the in-lab robot poses from recorded robot joint angles, yielding cleaner action annotations than the in-the-wild data despite their smaller scale.

\begin{table}[h]
\centering
\small
\begin{tabular}{lrrrrcr}
\toprule
Source & FPS & Hours & Frames & Raw \% & Upsamp. & Eff. \% \\
\midrule
EgoDex~\citep{egodex}                       & 30 & 826      & 89.2\,M  & 2.6  & 4.4$\times$  & 10.0 \\
In-lab human data         & 20 & 287      & 20.6\,M  & 0.6  & 5.7$\times$  & 3.0  \\
In-lab robot data         & 20 & 88       & 6.3\,M   & 0.2  & 12.6$\times$ & 2.0  \\
In-the-wild egocentric videos                        & 30 & 30{,}928 & 3.34\,B  & 96.6 & 1.0$\times$  & 85.0 \\
\midrule
\textbf{Total}               & -- & \textbf{32{,}129} & \textbf{3.46\,B} & 100 & -- & 100 \\
\bottomrule
\end{tabular}
\caption{Composition of the pre-training corpus. Effective sampling percentages are proportional to $\text{frames} \times \text{upsampling factor}$ and correspond to the distribution used during training.}
\label{tab:pretrain_data}
\end{table}

\minisection{Action Representation} Each training sample consists of an egocentric image paired with a future action chunk of horizon $T=64$. At each timestep, the action is represented by the poses of 42 hand joints (21 per hand). Each joint is encoded using a 9-dimensional representation comprising a 3D translation and a continuous 6D rotation representation~\citep{zhou2020continuityrotationrepresentationsneural}, resulting in 378 dimensions per timestep and 24{,}192 dimensions for the full action chunk.

Action targets are expressed as pose deltas relative to the current-frame hand pose and are normalized independently per dimension using mean and standard deviation statistics computed over the training set. A per-timestep validity mask is used to ignore padding near sequence boundaries when fewer than $T$ future steps are available.

Our decision to represent actions as a temporal chunk was inspired by ACT~\citep{zhao2023learningfinegrainedbimanualmanipulation}, which showed that predicting chunks of actions, rather than single-step actions, improves imitation learning by capturing temporal structure in demonstrations. We adapt this intuition to the pre-training setting, encouraging our vision encoder to produce latents that represent action over longer horizons.

\minisection{Preprocessing}
Input frames are decoded to RGB and resized to $256 \times 256$, matching the input resolution of the SigLIP 2 ViT-L/16 backbone. Pixel values are scaled to $[0,1]$ and normalized using the standard CLIP channel-wise mean $(0.481, 0.458, 0.408)$ and standard deviation $(0.269, 0.261, 0.276)$. During training, we apply a light random resized crop with scale sampled from $[0.9, 1.0]$, aspect-ratio jitter in $[0.75, 1.33]$, and bicubic interpolation. At inference, images are resized to a shorter side length of $256$ using bicubic interpolation and center-cropped to $256 \times 256$. Text inputs consist of the natural-language task instruction associated with each clip. Instructions are canonicalized by lowercasing and removing punctuation, then tokenized using the SigLIP 2 tokenizer with a maximum context length of $64$ tokens.

\subsection{Architecture}
\label{app:architecture}

Our model is an image-text-action contrastive architecture built on the SigLIP 2 backbone, with two modifications relative to the original design: (i) the vision tower's global attention-pooling head is replaced with a \emph{text-conditioned cross-attention pooling} module, and (ii) an \emph{action encoder} is introduced during pre-training. The action tower injects action supervision into the shared embedding space but is discarded after pre-training; only the image and text encoders are retained for downstream policy learning. We instantiate three backbone scales---ViT-B/16, ViT-L/16, and ViT-SO400M/16---which differ only in transformer width and depth. Architecture hyperparameters are summarized in ~\Cref{tab:arch_size} and~\Cref{tab:arch_shared}.

\minisection{Vision tower} The vision encoder is a SigLIP 2 Vision Transformer operating at $256\times256$ resolution with $16\times16$ patches, producing 256 patch tokens. We remove the default SigLIP MAP pooling head and expose the full patch-token sequence to a downstream pooling module. The ViT-B, ViT-L, and ViT-SO400M variants use widths of $768$, $1024$, and $1152$, depths of $12$, $24$, and $27$, and $12$, $16$, and $16$ attention heads, respectively. The SO400M variant follows the original SigLIP configuration with an MLP ratio of $3.7362$, while the base and large models use a ratio of $4$.

\minisection{Text tower} The text encoder is the corresponding SigLIP 2 text transformer with bidirectional self-attention. Inputs are tokenized using the SigLIP 2 multilingual (Gemma) tokenizer with a vocabulary of $256$k tokens and a context length of $64$. The encoder produces both a pooled text embedding and the full sequence of token representations. The pooled embedding is unused, and the token-level features are used to condition image pooling.

\minisection{Text-conditioned cross-attention pooling} We condition image pooling on the paired language instruction. Let $X \in \mathbb{R}^{B\times N\times C}$ denote the visual patch tokens and $T_L \in \mathbb{R}^{B\times L\times C}$ the text-token features. After applying the LayerNorm and linear projection, text tokens serve as queries while visual tokens provide keys and values in a multi-head cross-attention operation. This yields a sequence of text-conditioned visual features. The resulting sequence is then aggregated using a learned-query pooling operation consisting of a single trainable query token attending over the conditioned features, followed by LayerNorm and a linear projection. The final output is a text-conditioned image embedding $z_{\mathrm{img}}\in\mathbb{R}^{C}$. The cross-attention module uses $8$ heads for ViT-B and $16$ heads for ViT-L and SO400M, with attention dropout of $0.1$.

\minisection{Action encoder} To inject action information during pre-training, we introduce a lightweight Transformer encoder operating on future action sequences. Given a horizon-$T$ action chunk $a\in\mathbb{R}^{B\times T\times A_d}$, each action vector is projected into the shared embedding space, augmented with learned positional embeddings, and prepended with a learnable class token. The resulting sequence is processed by a $4$-layer Transformer encoder with $8$ attention heads, feed-forward dimension $4C$, GELU activations, and dropout $0.1$. The final class-token representation is projected to obtain the action embedding $z_{\mathrm{act}}\in\mathbb{R}^{C}$. For all our experiments, the action horizon is $T=64$ and the action dimensionality is $A_d=378$, corresponding to the full two-hand pose representation described in Section~\ref{app:pretraining_data}. Padded timesteps are masked during attention. The action encoder is trained jointly with the image and text towers and removed after pre-training.

\minisection{Initialization} The vision and text towers are initialized from publicly released SigLIP 2 checkpoints at the corresponding backbone scale. All newly introduced modules are randomly initialized. The contrastive temperature and bias follow the SigLIP parameterization, with the logit bias initialized to $-10$.

\begin{table}[t]
\centering
\small
\setlength{\tabcolsep}{8pt}
\renewcommand{\arraystretch}{1.15}
\caption{Architecture configuration that scales with vision encoder size. The vision and text towers share the same width, depth, and head count at each size. The cross-attention pooling module uses an embedding dimension matched to the tower width.}
\label{tab:arch_size}
\begin{tabular}{llccc}
\toprule
& & ViT-B/16 & ViT-L/16 & ViT-SO400M/16 \\
\midrule
\multirow{4}{*}{\textit{Vision tower}}
& Width      & 768 & 1024 & 1152 \\
& Depth      & 12  & 24   & 27 \\
& Heads      & 12  & 16   & 16 \\
& MLP ratio  & 4   & 4    & 3.74 \\
\midrule
\multirow{3}{*}{\textit{Text tower}}
& Width & 768 & 1024 & 1152 \\
& Depth & 12  & 24   & 27 \\
& Heads & 12  & 16   & 16 \\
\midrule
\multirow{2}{*}{\shortstack[l]{\textit{Cross-attn}\\\textit{pooling}}}
& Embed dim $C$ & 768 & 1024 & 1152 \\
& Heads           & 8   & 16   & 16 \\
\bottomrule
\end{tabular}
\end{table}

\begin{table}[t]
\centering
\small
\setlength{\tabcolsep}{8pt}
\renewcommand{\arraystretch}{1.15}
\caption{Architecture configuration shared across all vision encoder sizes.}
\label{tab:arch_shared}
\begin{tabular}{lll}
\toprule
Component & Parameter & Value \\
\midrule
\multirow{3}{*}{Text tower}
& Context length & 64 \\
& Vocab size     & 256{,}000 \\
& Attention      & Bidirectional \\
\midrule
Cross-attn pooling
& Query source & All text tokens \\
\midrule
\multirow{5}{*}{Action encoder}
& Layers / heads   & 4 / 8 \\
& FFN dim          & $4 C$ \\
& Horizon $T$      & 64 \\
& Action dim $A_d$ & 378 \\
\midrule
Shared & Dropout & 0.1 \\
\bottomrule
\end{tabular}
\end{table}

\subsection{Training}
\label{app:vision_training_details}

\minisection{Contrastive Image--Action Objective} Let $\mathbf{z}^{\text{img}}_i \in \mathbb{R}^C$ and $\mathbf{z}^{\text{act}}_i \in \mathbb{R}^C$ denote the L2-normalized text-conditioned image embedding and action embedding for the $i$-th sample in a batch of size $B$. For each pair $(i, j)$, we compute a similarity logit
\[
\ell_{ij} = t \cdot \langle \mathbf{z}^{\text{img}}_i, \mathbf{z}^{\text{act}}_j \rangle + b,
\]
where $t$ is a learnable temperature (parameterized in log-space and clamped at a maximum value) and $b$ is a learnable bias.

We define binary labels $y_{ij} \in \{+1, -1\}$ with $y_{ij} = +1$ for matching image--action pairs (i.e., $i=j$) and $y_{ij} = -1$ otherwise. The training objective is a full-batch sigmoid contrastive loss:
\[
\mathcal{L} = -\frac{1}{B} \sum_{i=1}^{B} \sum_{j=1}^{B} \log \sigma\!\left( y_{ij} \cdot \ell_{ij} \right),
\]
where $\sigma(\cdot)$ denotes the sigmoid function.

Positive pairs are constructed from the same trajectory: an image at time $t_0$ is paired with the subsequent action chunk spanning $t_0+1, \dots, t_0+T$, where $T$ is the horizon length. All other image--action pairs in the batch serve as negatives.

\minisection{Batch and masking} The loss is computed over the full global batch, where every image is contrasted against every action ($B = 32{,}768$ for our reference run). We do not apply explicit false-negative filtering; supervision is fully determined by the diagonal labeling structure above.

Two masking operations are applied upstream of the loss: (i) padded action frames in variable-length sequences are masked within the action encoder, and (ii) padded text tokens are masked in the text-conditioned cross-attention pooling module, ensuring that padding does not contribute to pooled representations.

\minisection{Optimization} We train with AdamW ($\beta_1{=}0.9$, $\beta_2{=}0.98$, $\epsilon{=}10^{-6}$, weight decay $1\times10^{-4}$), using a linear warmup of $2{,}000$ steps followed by cosine decay to zero. The temperature $t$ is initialized to $1/0.07 \approx 14.3$ and clamped at a maximum of $100$; in practice it saturates at this bound during training. The logit bias $b$ is initialized to $-10$. Actions are normalized per channel using dataset-level mean--std statistics.

We train for one full pass over the complete dataset with no gradient clipping. The vision and text towers are initialized from pretrained SigLIP\,2 weights, while the cross-attention pooling module and action encoder are trained from scratch (Appendix~\ref{app:architecture}).

\minisection{Hardware and throughput} The reference ViT-L/16 model is trained in \texttt{bf16} mixed precision (\texttt{amp\_bf16}) on 128 GPUs, using a per-GPU micro-batch of 128 and gradient accumulation of 2 for a global batch of 32,768.

\begin{table}[t]
\centering
\small
\caption{Training configuration for the ViT-L/16 model. The global batch is computed as per-GPU batch $\times$ number of GPUs $\times$ gradient accumulation factor, and provides in-batch negatives for the sigmoid contrastive loss.}
\label{tab:train}
\begin{tabular}{ll}
\toprule
Optimizer                & AdamW \\
$(\beta_1, \beta_2)$     & $(0.9,\ 0.98)$ \\
$\epsilon$               & $10^{-6}$ \\
Weight decay             & $1\times10^{-4}$ \\
Peak learning rate       & $1\times10^{-4}$ \\
LR schedule              & cosine decay to $0$ \\
Warmup                   & $2{,}000$ steps \\
Epochs                   & $1$ \\
Per-GPU micro-batch      & $128$ \\
Gradient accumulation    & $2$ \\
Global batch             & $32{,}768$ \\
Precision                & \texttt{bf16} AMP \\
Gradient clipping        & none \\
Temperature $t$ (init / max) & $1/0.07 \approx 14.3$ / $100$ \\
Logit bias $b$ (init)    & $-10$ \\
Action normalization     & per-channel mean--std \\
Tower init               & pretrained SigLIP\,2 \\
GPUs                     & $128$ H100 ($16$ nodes $\times\,8$) \\
\bottomrule
\end{tabular}
\end{table}

\section{Downstream Policy Training and Evaluation}
\label{app:downstream_policy_and_eval}
\subsection{Tasks}
\label{sec:tasks}

We evaluate on six real-world manipulation tasks chosen to span deformable-object
handling, dynamic and granular manipulation, multi-object sequencing, and
fine-grained dexterity. Each policy is trained from scratch on 200 teleoperated
demonstrations (150 for Pour Almonds) and evaluated over 12 trials. Five of the
six tasks are bimanual; only Turn On Lamp is single-arm.

\paragraph{Fold Shorts.}
\emph{Language instruction:} ``A pair of gray shorts lay on the tabletop. Grasping
them from underneath with both hands, fold them upward; then, using your right
hand to hold down the right side, use your left hand to grasp the left side and
fold them upward a second time.''
\par\smallskip\noindent
This bimanual task tests manipulation of a highly deformable object whose
configuration changes unpredictably under contact, and requires coordinated
two-handed control across a two-stage fold in which the second fold depends on
the result of the first.

\begin{center}
\includegraphics[width=0.9\linewidth]{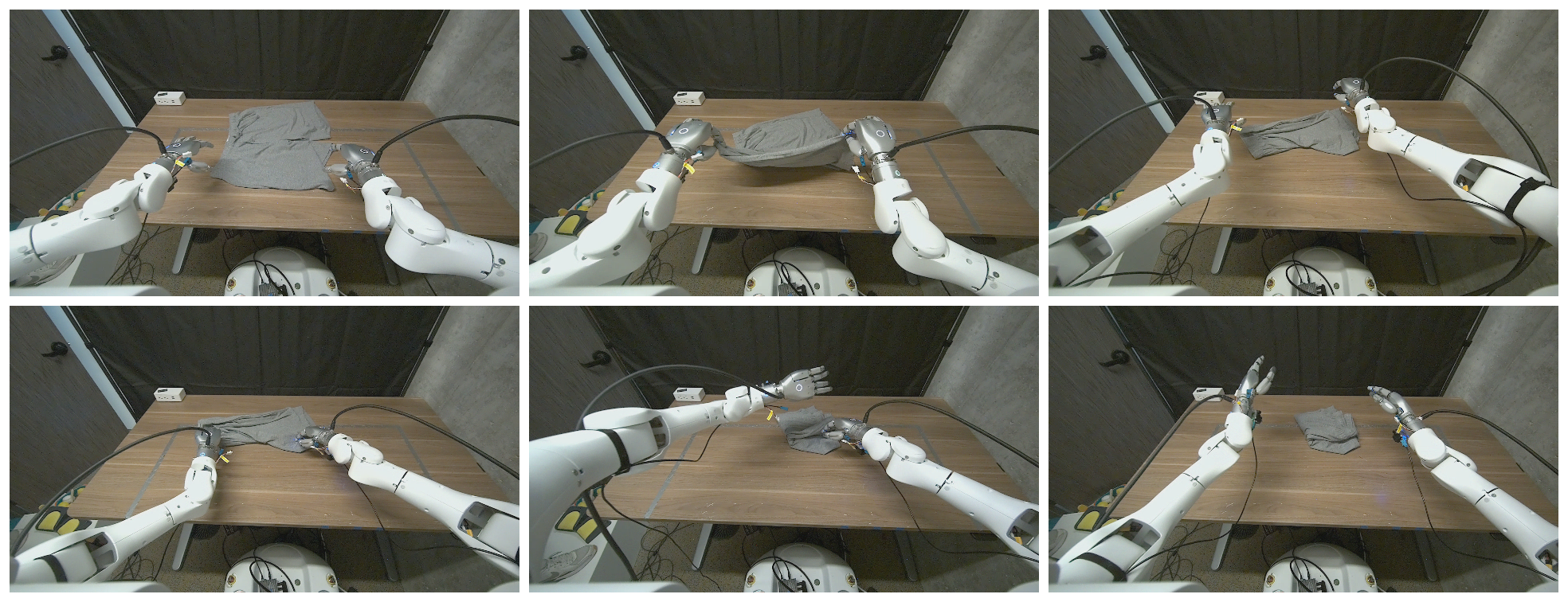}
\captionof{figure}{Progression of Fold Shorts.}
\label{fig:prog-fold-shorts}
\end{center}

\paragraph{Pour Almonds.}
\emph{Language instruction:} ``Pour the almonds from the filled cup to the empty
cup.''
\par\smallskip\noindent
This bimanual task probes control of a dynamic, granular process: the policy must
regulate cup orientation and pour rate to transfer free-flowing almonds without
spilling or overshooting. It is our most data-constrained task, trained on only
150 demonstrations.

\begin{center}
\includegraphics[width=0.9\linewidth]{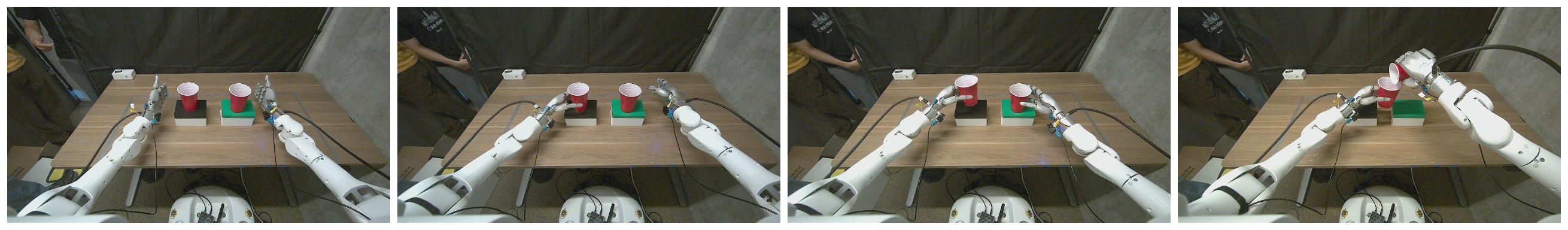}
\captionof{figure}{Progression of Pour Almonds.}
\label{fig:prog-pour-almonds}
\end{center}

\paragraph{Pick Fruits.}
\emph{Language instruction:} ``Pick up the fruit on the left side using your left
hand and place it in the basket. Then, pick up the fruit on the right side using
your right hand and place it in the basket.''
\par\smallskip\noindent
This bimanual task evaluates sequential pick-and-place over multiple objects,
testing reliable grasping of irregularly shaped items and correct hand--object
assignment across sub-goals.

\begin{center}
\includegraphics[width=0.9\linewidth]{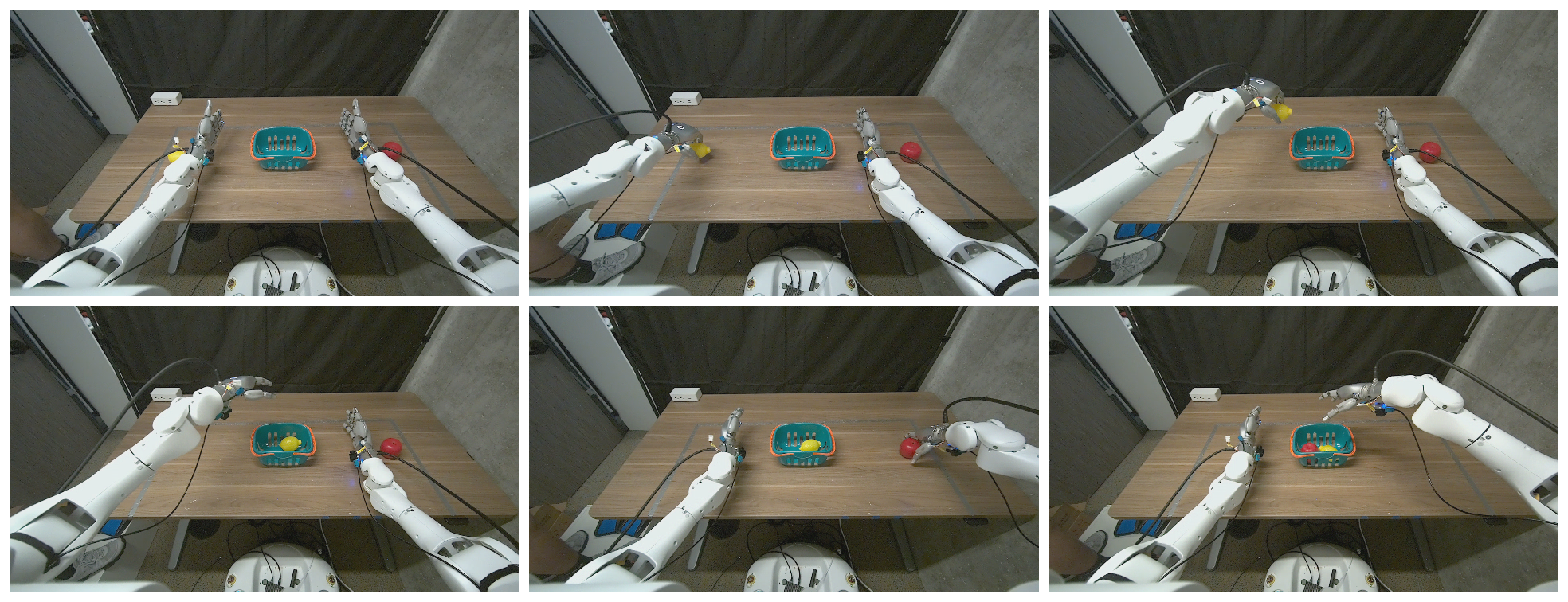}
\captionof{figure}{Progression of Pick Fruits.}
\label{fig:prog-pick-fruits}
\end{center}

\paragraph{Dispense Soap.}
\emph{Language instruction:} ``Use your left hand to pick up the soap dispenser,
and then use your right hand to press the pump to dispense soap into the red
bowl.''
\par\smallskip\noindent
This bimanual task requires asymmetric coordination in which one hand stabilizes
the dispenser while the other applies a controlled downward press, testing precise
force application against a compliant mechanism.

\begin{center}
\includegraphics[width=0.9\linewidth]{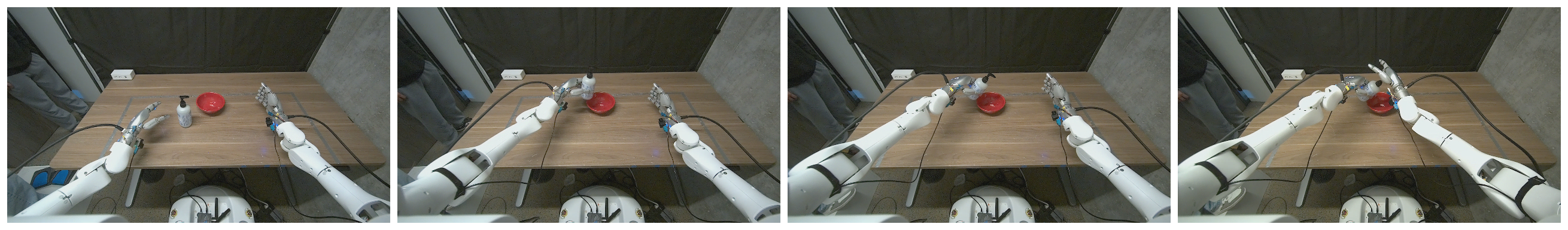}
\captionof{figure}{Progression of Dispense Soap.}
\label{fig:prog-dispense-soap}
\end{center}

\paragraph{Turn On Lamp.}
\emph{Language instruction:} ``Using your left hand, carefully pull the lamp chain
and release it to turn on the lamp.''
\par\smallskip\noindent
This single-arm task targets fine-grained dexterity: the policy must grasp a thin,
freely hanging chain, pull it through a short actuation stroke, and release,
leaving little margin for imprecise contact.

\begin{center}
\includegraphics[width=0.9\linewidth]{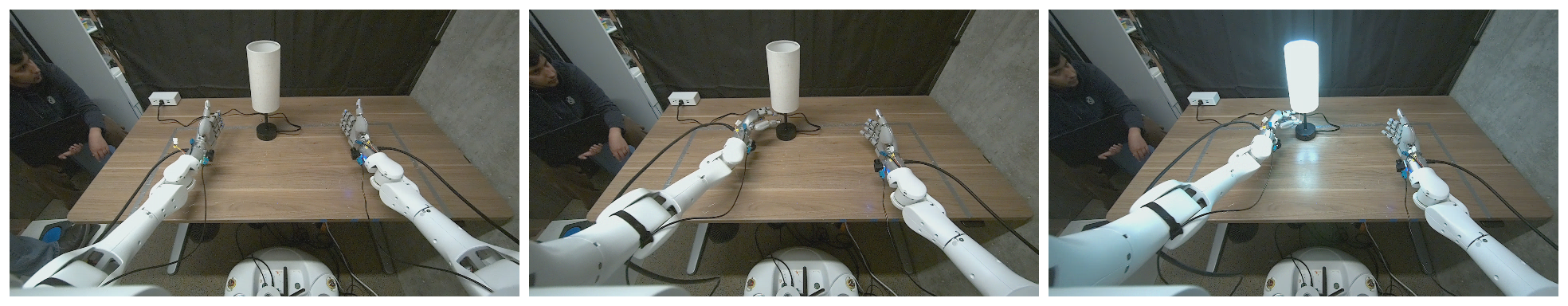}
\captionof{figure}{Progression of Turn On Lamp.}
\label{fig:prog-lamp}
\end{center}

\paragraph{Pull Tissue.}
\emph{Language instruction:} ``Use your left hand to pick up the tissue box, and
then use your right hand to pull out the tissue.''
\par\smallskip\noindent
This bimanual task tests coordinated extraction of a flexible object, where one
hand secures the box while the other gently pulls a single tissue free without
tearing it or dislodging the box.

\begin{center}
\includegraphics[width=0.9\linewidth]{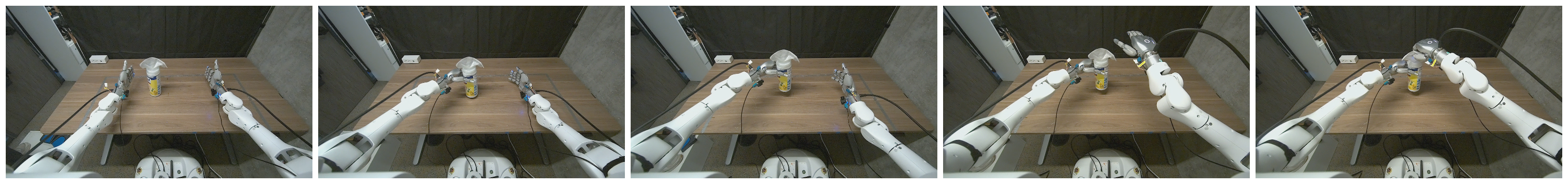}
\captionof{figure}{Progression of Pull Tissue.}
\label{fig:prog-pull-tissue}
\end{center}

\subsection{Policy Training}

We train a separate policy for each downstream task on top of the frozen vision encoder. The policy backbone is a Qwen3.5-0.8B~\citep{qwen3.5} decoder, trained from scratch. The vision encoder produces per-patch image tokens (or a single pooled/CLS token per image) which are linearly projected along with the text tokens into the policy's hidden dimension and concatenated with the flow-matching timestep embedding and noisy action tokens to form the policy input sequence. The action head uses flow matching, and we do not use proprioceptive or tactile inputs.

We train with AdamW (learning rate $10^{-4}$, no weight decay) and a cosine schedule without warmup. We use bfloat16 mixed precision and gradient clipping at $1.0$. Each task is trained for 200 epochs with a per-GPU batch size of 8 and no gradient accumulation. We train on 200 demonstrations per task, except for the almond-pouring task which uses 150. Each policy is trained on 4 nodes of 8 H100 GPUs (32 GPUs total, effective batch size 256), taking roughly 12--15 hours per task. Each baseline encoder uses its own native input preprocessing (image resolution and normalization); all other policy training settings are held fixed across encoders for a controlled comparison.

\subsection{Evaluation Protocol}
% - How success is measured
% - Rollouts per condition, initial state grid
% - Grading criteria

We evaluate each policy with 12 rollouts per task. For every task we
define a fixed grid of initial states that systematically varies the pose of
the manipulated objects across the reachable workspace; where object
orientation matters, the grid additionally samples a small set of rotations.
The same grid is used for all policies so that comparisons are matched on
initial conditions. All placements are referenced to fixed fiducials on the
table (the black background edge, the table edge, the hanger-mount markings,
and taped reference lines), making the grid reproducible across sessions.

Success is scored per rollout against a task-specific rubric. Single-stage
tasks use binary success, while multi-stage tasks award partial credit for
completing individual sub-stages (e.g., a successful grasp or alignment), with
full credit reserved for completing the entire motion.

\paragraph{Fold Shorts.} Initial states span a $3\times2$ grid of workspace
positions, each evaluated at two $\pm25^\circ$ tilts from neutral. We award
$+0.25$ per completed fold and full credit when both folds succeed.

\paragraph{Pick Fruits.} Fruit is placed at 2 positions on the left and 3 on
the right along the table edge, with the apple and lemon swapped across sides.
We award $+0.25$ for each fruit picked and placed in the basket, and full
credit when both are completed.

\paragraph{Pour Almonds.} The end and start cups are placed over a $3\times4$ grid of
positions relative to the hanger midline. We award $+0.25$ per cup picked and
full credit for a successful pour.

\paragraph{Dispense Soap.} The dispenser is placed over a $6\times2$ grid
between the reference lines on the left edge. We award $+0.25$ for picking the
dispenser and aligning it over the bowl, and full credit when soap is
successfully dispensed into the bowl.

\paragraph{Turn On Lamp.} We have a $4\times3$ grid for the lamp. Each is scored as binary success/fail.

\paragraph{Pull Tissue.} Initial states span a $4\times3$ grid. We award
$+0.25$ for grasping the tissue box and full credit for completing the full pull
motion.

\subsection{Environmental Robustness Experiment Setting}
\label{app:downstream_policy_and_eval:env-robustness}
All policies are trained only on demonstrations collected under the original,
clean conditions; the perturbations below are introduced solely at evaluation
time to probe the sensitivity of the learned visual representation.

\paragraph{Lighting.}
We consider two lighting perturbations relative to the original training
condition. In the \emph{Light} setting, we place an additional bulb above and in
front of the robot, casting extra shadows across the workspace. In the
\emph{Dark} setting, we reduce the intensity of the standard scene lighting. The
three conditions are shown side by side in Figure~\ref{fig:robust-lighting}.

\begin{center}
\includegraphics[width=\linewidth]{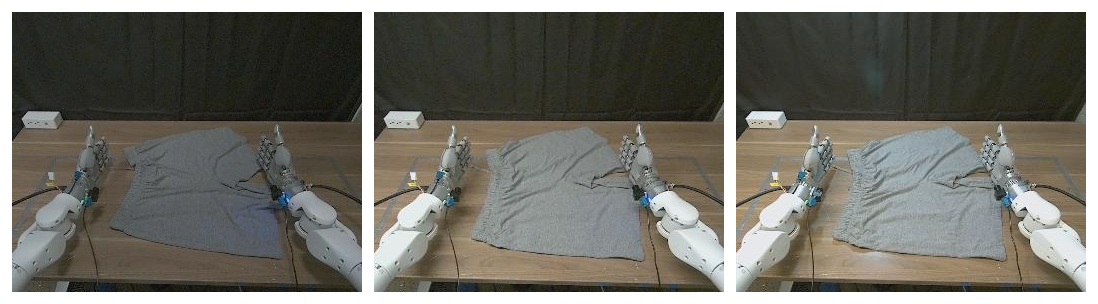}
\captionof{figure}{\textbf{(Left)} The \emph{Dark} condition reduces the intensity
of the standard scene lighting. \textbf{(Center)} The original lighting used
during training. \textbf{(Right)} The \emph{Light} condition adds an overhead bulb
that casts shadows across the workspace.}
\label{fig:robust-lighting}
\end{center}

\paragraph{Distractors.}
For the distractor setting, we add two objects to the scene: a red book and a
multi-colored Hanoi toy tower. Both are placed well within the manipulation area
so that they remain clearly visible in the egocentric camera view, as shown in
Figure~\ref{fig:robust-distractor}, and their positions are randomized across
trials.

\begin{center}
\includegraphics[width=\linewidth]{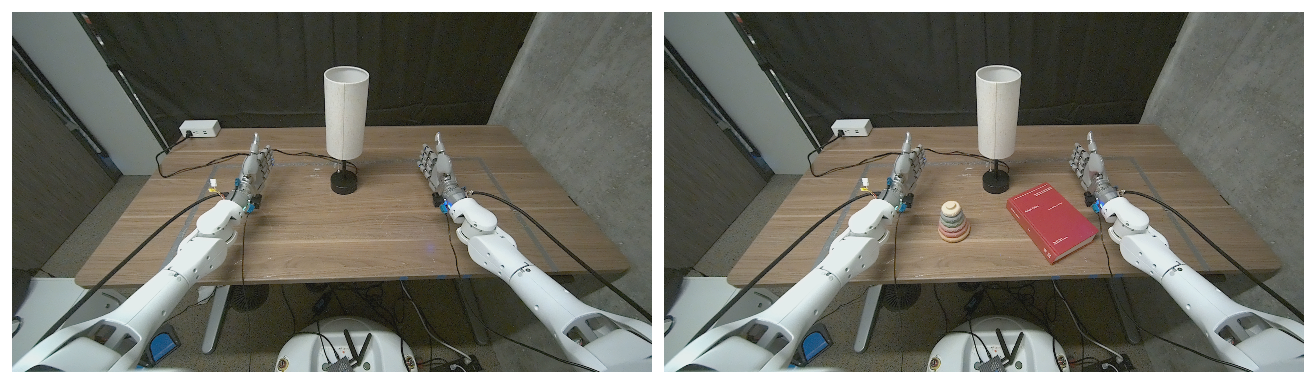}
\captionof{figure}{\textbf{(Left)} The original scene without distractors.
\textbf{(Right)} The scene with the two distractor objects added.}
\label{fig:robust-distractor}
\end{center}

\section{Baseline Encoders}
\label{app:baselines}

We evaluate the following baseline vision encoders, all frozen during policy training. The token strategy (CLS/pooled single token vs.\ all patch tokens) was selected per encoder based on which option yielded the best policy performance; see Table~\ref{tab:baseline_summary} for the selected strategy and other per-encoder specifications.

\minisection{Shared adaptation} All baselines are integrated into the policy through the same lightweight pipeline. The encoder's image and text tokens are each passed through a LayerNorm followed by a single linear projection (Xavier-initialized) that maps from the encoder's native output dimension into the policy's hidden dimension. Vision-only baselines (i.e., those without a paired text encoder) all use the CLIP ViT-L/14~\citep{radford2021learningtransferablevisualmodels} text encoder for the language stream, ensuring that differences in policy performance reflect differences in vision representations rather than text representations.

\minisection{SigLIP~\citep{zhai2023sigmoidlosslanguageimage}} We use \texttt{google/siglip-so400m-patch14-384}, with both vision and text towers loaded from the same checkpoint. Images are normalized with SigLIP's native statistics.

\minisection{SigLIP 2~\citep{tschannen2025siglip2multilingualvisionlanguage}} We use \texttt{google/siglip2-so400m-patch14-384}, with both vision and text towers loaded from the same checkpoint. Preprocessing matches SigLIP 2's native pipeline.

\minisection{DINOv2~\citep{oquab2024dinov2learningrobustvisual}} Vision tower: \texttt{facebook/dinov2-large}, a ViT-L/14 trained with self-supervised image objectives. Images are normalized with ImageNet statistics. We drop the CLS and register tokens and use the full patch grid.

\minisection{MVP~\citep{xiao2022maskedvisualpretrainingmotor}} Vision tower: ViT-L/16 pretrained with MAE on the datasets described in the original work. Images are bilinearly resized and normalized with ImageNet statistics.

\minisection{R3M~\citep{nair2022r3muniversalvisualrepresentation}} Vision tower: ResNet-50 pretrained with R3M's time-contrastive and video-language alignment objective. Images are resized to 256 along the shorter side, center-cropped, and normalized with ImageNet statistics. We use R3M's global pooled feature as the single token.

\minisection{VC-1~\citep{majumdar2023searchartificialvisualcortex}} Vision tower: ViT-L/16 pretrained with MAE on the egocentric video and ImageNet mixture described in~\citep{majumdar2023searchartificialvisualcortex}. Images are bicubically resized to 256 along the shorter side, center-cropped, and normalized with ImageNet statistics.

\minisection{VideoMAE~\citep{tong2022videomaemaskedautoencodersdataefficient}} Vision tower: \texttt{MCG-NJU/videomae-large}, a ViT-L with 3D patch embedding (temporal tubelet $=2$, spatial patch $=16$). At each policy step, we stack two consecutive frames from the same camera (the current frame and the frame at $t -2$, clamped at the trajectory boundary) to form the 3D input. Images are bilinearly resized and normalized with ImageNet statistics.

\minisection{Qwen3.5 Vision Encoder~\citep{qwen3.5}} As an additional reference point, we evaluate the native Qwen3.5-0.8B vision tower loaded from the same checkpoint used for the policy backbone. Unlike the other baselines, this encoder shares the policy's hidden dimension natively, so the LayerNorm and linear projection layers are replaced with identity mappings. Text bypasses the latent stage entirely and is embedded by the policy backbone's native token embeddings.

\begin{table}[h]
\centering
\small
\begin{tabular}{lccccc}
\toprule
Encoder & Vision arch. & Image res. & Token strategy & Vision dim & Text tower \\
\midrule
SigLIP        & ViT-SO400M/14 & $384$         & patch & 1152 & SigLIP \\
SigLIP 2      & ViT-SO400M/14 & $384$         & patch & 1152 & SigLIP 2 \\
DINOv2        & ViT-L/14      & $224$         & patch & 1024 & CLIP ViT-L/14 \\
MVP           & ViT-L/16      & $256$         & cls   & 1024 & CLIP ViT-L/14 \\
R3M           & ResNet-50     & $224$         & cls   & 2048 & CLIP ViT-L/14 \\
VC-1          & ViT-L/16      & $224$         & cls   & 1024 & CLIP ViT-L/14 \\
VideoMAE      & ViT-L/16 (3D) & $224 \times 2$ frames & patch & 1024 & CLIP ViT-L/14 \\
Qwen   & Qwen3.5 native ViT & $384 \times 288$ & patch & 1024 & shared with LM \\
\bottomrule
\end{tabular}
\caption{Summary of baseline vision encoders. ``Token strategy'' indicates whether we feed a single CLS/pooled feature or the full per-patch grid into the policy. The token strategy was selected per encoder based on downstream performance.}
\label{tab:baseline_summary}
\end{table}

\minisection{Direct Regression Baseline} We also experimented with direct action regression as a pre-training objective, using both (i) an MLP head operating on the pooled text-conditioned image embedding and (ii) a transformer decoder operating on the full ViT patch sequence. Both variants were initialized from the same SigLIP 2 checkpoint and trained with the identical dataset, optimizer, and schedule as our contrastive model. Training supervised a chunk of future actions using an $L_1$ loss on per-dimension-normalized targets, masked over valid timesteps; we additionally evaluated MSE and Huber losses, which yielded similar results.

The MLP variant regresses the entire action chunk from the single pooled image-text embedding. The decoder variant instead discards the SigLIP cross-attention pooling head and applies a 4-layer transformer decoder (hidden size $1024$, $16$ attention heads, MLP ratio $4$). A set of learned per-timestep query tokens, each conditioned by the text embedding, cross-attends to the full ViT patch sequence before a shared linear projection maps decoder outputs to the action space.

Neither variant produced useful representations. In both cases, training failed to learn a useful mapping: the regression loss plateaued early and the resulting features carried no meaningful structure for downstream evaluation. These results motivated our use of a contrastive objective in all main experiments.
\end{document}